\documentclass[twoside]{article}

\usepackage{aistats2020}
\usepackage{amssymb}
\usepackage{amsmath}
\usepackage{amsfonts}
\usepackage{graphicx}
\usepackage{epstopdf,epsf,psfrag}
\usepackage{bbm}
\usepackage{dsfont}

\usepackage[utf8]{inputenc}
\usepackage{colortbl}
\usepackage{hyperref}
\usepackage{framed}
\usepackage{multirow}

\usepackage[linesnumbered, english]{algorithm2e}
\usepackage{natbib}
\hypersetup{
    colorlinks,
    linkcolor={black},
    citecolor={blue},
    urlcolor={blue}
}
\newtheorem{definition}{Definition}[section]
\newtheorem{theorem}{Theorem}[section]
\newtheorem{proposition}{Proposition}[section]
\newtheorem{lemma}{Lemma}[subsection]
\begin{document}

%

%

\twocolumn[

\aistatstitle{The Area of the Convex Hull of Sampled Curves:\\
a Robust Functional Statistical Depth Measure}

\aistatsauthor{ Guillaume Staerman \And Pavlo Mozharovskyi \And  Stephan Cl\'{e}men\c{c}on }

\aistatsaddress{ LTCI, T\'{e}l\'{e}com Paris, Institut Polytechnique de Paris } ]

\begin{abstract}
With the ubiquity of sensors in the IoT era, statistical observations are becoming increasingly available in the form of massive (multivariate) time-series. 
Formulated as unsupervised anomaly detection tasks, an abundance of applications like aviation safety management, the health monitoring of complex infrastructures or fraud detection can now rely on such functional data, acquired and stored with an ever finer granularity.
The concept of \textit{statistical depth}, which reflects centrality of an arbitrary observation w.r.t. a statistical population may play a crucial role in this regard, anomalies corresponding to observations with 'small' depth. Supported by sound theoretical and computational developments in the recent decades, it has proven to be extremely useful, in particular in functional spaces. However, most approaches documented in the literature consist in evaluating independently the centrality of each point forming the time series and consequently exhibit a certain insensitivity to possible shape changes. In this paper, we propose a novel notion of functional depth based on the area of the convex hull of sampled curves, capturing gradual departures from centrality, even beyond the envelope of the data, in a natural fashion. We discuss practical relevance of commonly imposed axioms on functional depths and investigate which of them are satisfied by the notion of depth we promote here. Estimation and computational issues are also addressed and various numerical experiments provide empirical evidence of the relevance of the approach proposed.
\end{abstract}


\section{Introduction}\label{intro}

Technological advances in data acquisition, management and warehousing (\textit{e.g.} IoT, distributed platforms) enable massive data processing and are leading to a wide variety of new applications in the digitalized (service) industry. The need to design more and more automated systems fed by ever more informative streams of data manifests in many areas of human activity (\textit{e.g} transportation, energy, health, commerce, finance/insurance). Monitoring the behavior/health of complex systems offers a broad spectrum of machine-learning implementation as classification or anomaly detection.
With the increasing industrial digitalization, data are more and more often collected in quasi-real time and naturally take the form of temporal series or functions. The case of functional data is thus of crucial interest in practice, refer to \textit{e.g.} \cite{ramsey2,ramsey} for an excellent account of Functional Data Analysis (FDA in abbreviated form) and of its applications. 
A functional dataset is typically a set of $n\geq 1$ curves partially observed at different time points $T_1=t_1<\ldots< t_p=T_2$ which can be seen as $n$ (partially observed)  realizations of a stochastic process $X=(X_t)_{t\in [T_1,T_2]}$. Hence, the first step of FDA generally consists in reconstruct the functional objects from these observations, by means of interpolation, smoothing or projection techniques. Whereas, with the design of many successful algorithms such as (deep) neural networks, SVM's or boosting procedures, the practice of statistical learning  has rapidly generalized these last few years, the automatic analysis of functional data to achieve complex (\textit{e.g.} unsupervised) tasks such as anomaly detection is still a challenge, due to the huge variety of possible patterns that may carry the relevant information for discrimination purposes. It is far from straightforward to generalize directly methods originally introduced in the finite-dimensional case to the functional setup, unless preliminary \textit{filtering} or dimensionality reduction techniques are used, see \textit{e.g}  \cite{SVMF,FIF}. Such techniques essentially consist in projecting the observation, supposed to take their values in a certain Hilbert space, onto a subspace of finite dimensionality, generally defined by truncating their expansion in a Hilbertian basis of reference or by means of a flexible dictionary of functions/'atoms'. Next, one can apply any state-of-the-art algorithm tailored to the finite dimension case, based on the parsimonious representations thus obtained, \textit{cf} \cite{ferraty,ramsey2}. The basis functions can be either chosen among a pre-selected dictionary (\textit{e.g.} Fourier, wavelets, cosine packets, \textit{etc.}) presumably capable of capturing the information carried by the curves or built empirically using Functional Principal Component Analysis, retaining the most informative part of the (Kahrunen-Loève) decomposition only, see \cite{ramsey}. Of course, the major drawback of such FDA approaches lies in the fact that they are dramatically conditioned by the finite-dimensional representation method chosen, the subsequent analysis of the data may be fully jeopardized if the latter makes disappear some patterns relevant for the task considered.
 
 Originally introduced by J. Tukey to extend the notion of median/quantile to multivariate random variables (see \cite{tukey}), a \textit{data depth} is a function defined on the feature space and valued in $[0,1]$ used to determine the 'representativeness' of a point with respect to a statistical population and that should fulfill a variety of desirable properties (ideally just like the quantile function in the univariate situation). Given a training set, a data depth function provides a score that measures the centrality of any element w.r.t. a dataset and then defines, notably, an ordering of all the elements of the dataset. In particular, it finds a very natural application in (unsupervised) anomaly detection, see \textit{e.g} \cite{cuevas,long5,mozha} in supervised situations and \cite{rousseeuw,nagy} in the unsupervised context: an observation is considered all the more 'abnormal', as its depth is small. This concept has been extended to the functional data framework by integrating univariate depth functions on the whole interval of time $[T_1,T_2]$, see \textit{e.g.} \cite{claeskens,Fraiman2001,rousseeuw}. Alternatively, various depth functions fully tailored to the functional setup have been introduced in the statistical literature,  refer to \textit{e.g} \cite{chak2,dutta,lopez,lopez2}. However, most of them fail to fulfill certain desirable properties or face significant computational difficulties.
It is the major purpose of this paper to introduce a novel robust functional statistical depth measure dedicated to the analysis of functional data. Based on the area of the convex hull of collections of sampled curves, it is easy to compute and to interpret both at the same time. Given a curve $x$ lying in a certain functional space $\mathcal{F}$ (\textit{e.g.} the Kolmogorov space $\mathcal{C}([0,1])$, the space of real valued continuous functions on $[0,1]$), the general idea is to quantify its contribution, on average, to the area of the convex hull (ACH in short) of random curves in $\mathcal{F}$ with the same probability law. Precisely, this function, referred to as the \textit{ACH depth} throughout the article, is defined by the ratio of the ACH of the sample to that of the sample augmented by the curve $x$. We prove here that it fulfills various properties desirable for depth functions. In particular, given its form, it exhibits \textit{sensitivity} (\textit{i.e.} the depth score of new/test curves that are further and further away from the training set of curves decreases smoothly), which property, quite desirable intuitively, is actually not satisfied by most statistical (functional) depth documented in the literature. For instance, the statistical Tukey depth assigns a score of $0$ to any element lying outside the convex hull of the training data, see \cite{Tukey75, dutta}. In addition, the statistical depth we promote here is robust to outliers: adding outliers to the training set of curves has little or no effect on the returned score and ordering on a test set. For this reason, this functional depth is very well suited for unsupervised anomaly detection. In the functional setup, this task is extremely challenging. Indeed, the richness of functional spaces leads to a huge diversity in the nature of possibly observed differences between curves. As dicussed in \cite{rousseeuw}, three main types of anomaly can be distinguished: \textit{shift}, \textit{magnitude} or \textit{shape} anomalies. Anomalies can be either \textit{isolated/transient} or \textit{persistent} depending of their duration ; some of them being more difficult to detect (\textit{shape} anomalies). Since the functional statistical depth measure we propose is related to a whole batch of curves and do not reflect the individual properties of a single curve, it enables the detection of a wide variety of  anomaly shapes, as illustrated by the numerical experiments displayed in Section ~\ref{section4}.  
 
 The paper is organized as follows. In Section ~\ref{section2}, basic concepts pertaining to the statistical depth theory, both for the multivariate framework and for the functional case, are briefly recalled for clarity's sake. In Section ~\ref{section3} the functional statistical depth based on the area of the convex hull of a batch of curves is introduced at length and its theoretical properties are investigated, together with computational aspects.  Section ~\ref{section4} presents numerical results in order to provide strong empirical evidence of the relevance of the novel depth function proposed, for the purpose of unsupervised functional anomaly detection especially. Eventually, concluding remarks are collected in Section ~\ref{section5}.

\section{Background and Preliminaries}\label{section2}

For clarity, we start with recalling the concept of \textit{statistical depth} in the multivariate and functional framework. We next list the desirable properties it should fulfill and also briefly review recent advances in this field. Here and throughout, the Dirac mass at any point $a$ is denoted by $\delta_a$, the convex hull of any subset $A$ of $\mathbb{R}^d$ by $conv(A)$.

\subsection{Data Depth in a Multivariate Space} \label{DD}

By \textit{data depth}, one usually means a nonparametric statistical function that determines the centrality of any element $x\in \mathbb{R}^d$ with respect to a statistical population. Given a dataset, a depth function provides a center-outward ordering of the data points. Since it permits to define a ranking of the (multivariate) observations and \textit{local averages} derived from it, a \textit{data depth} can be used for various tasks, including classification (\cite{LangeMM14}), clustering (\cite{jornsten}), anomaly detection (\cite{ser2006}) or rank tests (\cite{oja}). In order to give a precise definition, some notations are needed. Let $X$ be a random variable, defined on a certain probability space $(\Omega, \mathcal{A}, \mathbb{P} )$, taking its values in $\mathcal{X}\subset\mathbb{R}^d$ with probability distribution $P$. Denote by $\mathcal{P}(\mathcal{X})$ the set of all probability distributions on $\mathcal{X}$. A data depth is a function
$$
\begin{tabular}{cccc}
$D:$ & $\mathbb{R}^d\times \mathcal{P}(\mathcal{X})$ & $\longrightarrow$ & $[0,1]$
\\ &  $( x, \; P)$ & $\longmapsto$ & $D(x,\; P)$
\end{tabular}
$$
measurable with respect to its first argument $x$. It is interpreted as follows: the closer the quantity $D(x,P)$ to $1$, the deeper (\textit{i.e.} the more 'central')  the observation $x$ with respect to the distribution $P$ is considered. As mentioned above, it naturally defines a predorder on the set $\mathcal{X}$. In particular, medians of the multivariate distribution $P$ corresponds to maximizers of the depth function $D$ and quantile regions are defined as depth sublevel sets. A crucial example is the \textit{half-space} depth $D_T$ (also called \textit{location} depth sometimes) introduced in the seminal contribution \cite{Tukey75}. It is defined as
$$
D_T(x,\; P)=\inf\{ P(H):\; H \text{ closed half-space, } x\in H \},
$$
for any $x\in \mathbb{R}^d$ and probability distribution on $\mathbb{R}^d$.  As the distribution $P$ is generally unknown, a statistical version can be built from independent copies $X_1,\; \ldots,\; X_n$ of the generic random vector $X$ by means of the \textit{plug-in} principle, \textit{i.e.} by replacing $P$ by an empirical counterpart $\widehat{P}_n$, typically the raw empirical distribution $(1/n)\sum_{i=1}^n\delta_{X_i}$ (or a smooth/penalized version of the latter), yielding the empirical depth
\begin{equation}\label{eq:emp_depth}
\widehat{D}_n(x)=D(x,\; \widehat{P}_n).
\end{equation}
Empirical medians and quantile regions are then naturally defined as medians and quantile regions of the empirical depth \eqref{eq:emp_depth}. Of course, the relevance of a depth function regarding the measurement of \textit{centrality} in a multivariate space is guaranteed in the sole case where certain desirable properties are satisfied. 
  We refer to \cite{ZuoSerfling00} for an account of the statistical theory of multivariate data depth and many examples. 

\subsection{Statistical Functional Depth}\label{subsec:SFD}

In this paper, we consider the situation where the r.v. $X$ takes its values in a space of infinite dimension. Precisely, focus is on the case where the feature space is the vector space $ \mathcal{C}([0,1])$ of real-valued continuous functions on $[0,1]$:
\begin{equation*}\begin{array}{ccccc}
X& : &(\Omega,\; \mathcal{A},\; \mathbb{P}) &\longrightarrow&  \mathcal{C}([0,1]) \\
&& \omega &\longmapsto & X(\omega)=(X_t(\omega))_{t\in [0,1]}.
\end{array}
\end{equation*} Recall that, when equipped with the $\sup$ norm $||.||_{\infty}$,  $ \mathcal{C}([0,1])$ is a separable Banach space. 
We denote by $\mathcal{P}(\mathcal{C}([0,1]))$  the set of all probability laws on $\mathcal{C}([0,1])$ and by $P_{t}$ the $1$-dimensional marginal of the law $P$ of the stochastic process $X$ at time point $t\in [0,1]$.
 
Depths in a functional framework have been first considered in \cite{Fraiman2001}, where it is proposed to define functional depths as simple integrals over time of a univariate depth function $D$, namely $(x,P)\in \mathcal{C}([0,1])\mapsto \int_{0}^1 D(x_t,P_{t})dt$. Due to the averaging effect, local changes for the curve $x$ only induce slight modifications of the depth value, which makes anomaly detection approaches based on such `poorly sensitive' functional depths ill-suited in general. Recently, alternative functional depths have been introduced, see \cite{lopez,lopez2} for depths based on the geometry of the set of curves, \cite{chak2} for a notion of depth based on the $L_2$ distance or \cite{dutta} for a functional version of the Tukey depth. As discussed in \cite{nieto} and \cite{Gijbels}, the axiomatic framework introduced in \cite{ZuoSerfling00} for multivariate depth is no longer adapted to the richness of the topological structure of functional spaces. Indeed, the vast majority of the functional depths documented in the literature do not fulfill versions of the most natural and elementary properties required for a depth function in a multivariate setup, \textit{cf} \cite{Gijbels}.  However, there is still no consensus about the set of desirable properties that a functional depth should satisfy, beyond the form of sensitivity mentioned above. Those that appear to be the most relevant in our opinion are listed below. By $P_X$ is meant the law of a functional r.v. taking its values in $\mathcal{C}([0,1])$.

\begin{itemize}
\item[•](\textsc{Non-degeneracy}) For all non atomic distribution $P$ in $\mathcal{P}(\mathcal{C}([0,1]))$, we have $$\underset{x\in \mathcal{C}([0,1]) }{\inf} D(x ,P)<\underset{x\in \mathcal{C}([0,1])}{\sup} D(x ,P).$$
\item[•] (\textsc{Affine invariance}) The depth $D$ is said to be (scalar) affine invariant if for any $x$ in $\mathcal{C}([0,1]) $ and all $a$, $b$ in $\mathbb{R}$, we have $$D(x,P_X)=D(ax+b,P_{aX+b}).$$ 
\item[•] (\textsc{Maximality at the center}) For any point-symmetric and non atomic distribution $P$ with $\theta \in \mathcal{C}([0,1])$ as center of symmetry, we have 
$$
D(\theta, P)=\underset{x\in \mathcal{C}([0,1])}{\sup} D(x,P).$$  
\item[•] (\textsc{Vanishing at $\infty$}) For any non atomic distribution $P$ in $\mathcal{P}(\mathcal{C}([0,1]))$,
$$D(z , P_X )\underset{||z||_{\infty}\longrightarrow \infty}{\; \; \; \; \; \longrightarrow} \underset{x \in \mathcal{C}([0,1])}{\inf } D(x , P).$$
\item[•] (\textsc{Decreasing w.r.t. the deepest point}) For any 
$P$ in $\mathcal{P}(\mathcal{C}([0,1]))$  such that $D(z,P)=\underset{x \in \mathcal{C}([0,1])}{\sup} D(x,P)$, $D(x,P)<D(y,P)<D(z,P)$ holds for any $x,y \in \mathcal{C}([0,1])$ such that $\min \{d(y,z), d(y,x) \}>0$ and $\max \{d(y,z), d(y,x) \} < d(x,z)$.
\item[•] (\textsc{Continuity in $x$}) For any non atomic distribution $P \in \mathcal{P}(\mathcal{C}([0,1]))$,  the function $x\mapsto D(x, P)$ is continuous w.r.t. the $\sup$ norm.
\item[•] (\textsc{(Uniform-) continuity in $P$}) For all $x$ in $\mathcal{C}([0,1])$, the mapping $P\in \mathcal{P}(\mathcal{C}([0,1])) \mapsto D(x,\; P)$ is (uniformly-) continuous w.r.t. the L\'evy-Prohorov metric. 
\end{itemize}

Before introducing the ACH depth and investigating its properties, a few remarks are in order. Though it obviously appears as mandatory to make the other properties meaningful, \textit{non-degeneracy}, is actually not fulfilled by all the functional depths proposed, see \textit{e.g} \cite{lopez,lopez2,dutta}. The '\textit{Maximality at center}' and '\textit{Decreasing w.r.t. the deepest point}' properties permit to preserve the original center-outward ordering goal of \textit{data depth} in the functional framework. Many definition of the concept of "symmetry" in a functional space are detailed in the Supplementary material for the sake of place. The '\textit{Continuity in $x$}' property extends a property fulfilled by cumulative distribution functions of multivariate continuous distributions.
From a statistical perspective, the '\textit{Continuity in $P$}' property is essential, insofar as $P$ must be replaced in practice by an estimator, \textit{cf} Eq. \eqref{eq:emp_depth}, built from finite-dimensional observations, \textit{i.e.} a finite number of sampled curves. 

\section{The Area of the Convex Hull of (Sampled) Curves}\label{section3}

It is the purpose of this section to present at length the statistical depth function we propose for path-valued random variables. As shall be seen below, its definition is based on very simple geometrical ideas and various desirable properties can be easily checked from it. Statistical and computational issues are also discussed at length.
 By $\mathcal{K}_2$ is meant the collection of all compact subsets of $\mathbb{R}^2$ and $\lambda$ denotes Lebesgue measure on the plane $\mathbb{R}^2$. Consider an i.i.d. sample $X_1,\; \ldots,\; X_n$ drawn from $P$ in $\mathcal{P}(\mathcal{C}([0,1]))$.
  The graph of any function $x$ in $\mathcal{C}([0,1])$ is denoted by
\begin{equation*}\label{eqn:graphOne}
graph(x)= \{ (t,y): y=x(t), t \in [0,1] \},
\end{equation*}
while we denote by $graph(\{x_1,\ldots , x_n \})$ the set
\begin{equation*}\label{eqn:graph}
\bigcup_{i=1}^n graph(\{x_i\})
\end{equation*}
defined by a collection of $n\geq 1$ functions $\{x_1,\ldots , x_n \}$ in $\mathcal{C}([0,1])$.
We now give a precise definition of the statistical depth measure we propose for random variables valued in $\mathcal{C}([0,1])$.

 \begin{definition}\label{def:ACH} Let $ J \geq 1$ be a fixed integer. The ACH depth of degree $J$ is the function $D_{J}: \mathcal{C}([0,1]) \times \mathcal{P}(\mathcal{C}([0,1]))\rightarrow [0,1]$ defined by: $\forall x\in \mathcal{C}([0,1])$,
 
 \begin{equation*}\label{eqn:defPop}
D_{J}(x,\; P)=\mathbb{E}\left[  \frac{\lambda \left(  conv \left( graph \left(\{X_1, \ldots ,X_J \} \right) \right) \right) }{\lambda \left( conv \left( graph \left(\{X_1,\ldots ,X_J \} \cup \{x\} \right)\right) \right)} \right],
\end{equation*}
where $X_1,\; \ldots ,\; X_J$ are i.i.d. r.v.'s drawn from $P$.
Its average version $\bar{D}_{J}$ is defined by: $\forall x\in \mathcal{C}([0,1])$,
\begin{equation*}\label{eqn:defPopave}
\bar{D}_{J}(x,\; P)=\frac{1}{J}\sum_{j=1}^{J} D_{j}(x,\; P).
\end{equation*}
\end{definition}

The choice of $J$ leads to various views of distribution $P$, the average variant permitting to combine all of them (up to degree $J$). When $n\geq J$, an unbiased statistical estimation of $D_J(.,\; P)$ can be obtained by computing the symmetric $U$-statistic of degree $J$, see \cite{Lee90}: $\forall x\in \mathcal{C}([0,1])$,
\begin{multline}\label{eqn:defEmp}
D_{J,n}(x)=\\
\frac{1}{\binom{n}{J}} \sum_{1\leq i_1 < \ldots < i_J \leq n} \frac{\lambda \left( conv \left( graph \left(\{X_{i_1},\ldots ,X_{i_J} \} \right) \right) \right) }{\lambda \left( conv \left( graph \left(\{X_{i_1}, \ldots ,X_{i_J}, x\}\right) \right) \right)}.
 \end{multline}
 Considering the empirical average version given by
\begin{equation*}\label{eqn:defPopave}
\forall x\in \mathcal{C}([0,1]),\;\; \bar{D}_{J,n}(x)=\frac{1}{J}\sum_{j=1}^{J} D_{j,n}(x)
\end{equation*}
 brings some 'stability'. However, the computational cost rapidly increasing with $J$, small values of $J$ are preferred in practice. Moreover, as we illustrate in Section ~\ref{subsec:AS}, J equal two already yields satisfactory results. 

\noindent {\bf Approximation from sampled curves.} In general, one does not observes the batch of continuous curves $\{X_1,\; \ldots,\; X_n \}$ on the whole time interval $[0,1]$ but at discrete time points only, the number $p\geq 1$ of time points and the time points $ 0 \leq t_1<t_2< \ldots <t_p\leq 1$ themselves possibly varying depending on the curve considered. In such a case, the estimators above are computed from continuous curves reconstructed from the sampled curves available by means of interpolation procedures or approximation schemes based on appropriate basis. In practice, linear interpolation is used for this purpose with theoretical guarantees (refer to Theorem ~\ref{Approx} below) facilitating significantly the computation of the empirical ACH depth, see subsection ~\ref{subsec:AD}.

\subsection{Main Properties of the ACH Depth}\label{subsec:TP}
In this subsection, we study theoretical properties of the population version of the functional depths introduced above and next establish the consistency of their statistical versions. The following result reveals that, among the properties listed in the previous subsection, five are fulfilled by the (average) ACH depth function.

\begin{proposition}\label{Prop} For all $J\geq 1$, the depth function $D_{J}$ (respectively, $\bar{D}_{J}$) fulfills the following properties: '\textit{non-degeneracy}',  '\textit{affine invariance}',  '\textit{vanishing at infinity}',  '\textit{continuity in $x$}' and  '\textit{continuity in $P$}'. In addition, the following properties are not satisfied: '\textit{maximality at center}' and '\textit{decreasing w.r.t. the deepest point}'. 
\end{proposition}
Refer to the Appendix section for the technical proof. In a functional space, not satisfying \textit{maximality at center} is not an issue. For instance, though the constant trajectory $y(t)\equiv 0$ is a center of symmetry for the Brownian motion, it is clearly not representative of this distribution. In contrast, \textit{scalar-affine invariance} is relevant, insofar as it allows z-normalization of the functional data and 
\textit{continuity in $P$} is essential to derive the consistency of $D_{J,n}$ (respectively, of $\bar{D}_{J,n}$),  as stated below. 


\begin{theorem}\label{Asymp}
Let $J\geq 1$ and $X_1,\; \ldots,\; X_n$ be $n\geq J$ independent copies of a generic r.v. $X$ with distribution $P\in \mathcal{P}(\mathcal{C}([0,1]))$. As $n\to \infty$, we have, for any $x\in \mathcal{C}([0,1])$, with probability one,
$$\left\vert D_{J,n}(x)-D_{J}(x,\; P)\right\vert \rightarrow 0$$
and 
$$\left\vert \bar{D}_{J,n}(x)-\bar{D}_{J}(x,\; P) \right\vert \rightarrow 0.$$
\end{theorem}
\subsection{On Statistical/Computational Issues}\label{subsec:CA}

As mentioned above, only sampled curves are available in practice. Each random curve $X_i$ being observed at fixed time points $0=t^{(i)}_1<t^{(i)}_2<\ldots < t^{(i)}_{p_i}=1$ (potentially different for each $X_i$) with $p_i\geq 1$, we denoted by $X'_1,\; \ldots,\; X'_n$ the continuous curves reconstructed from the sampled curves $(X_i(t^{(i)}_1),\; \ldots,\; X_{i}(t^{(i)}_{p_i}))$, $1\leq i \leq n$, by linear interpolation.
From a practical perspective, one considers the estimator $D'_{J,n}(x)$ of $D_{J}(x,\; P)$ given by the approximation of $D_{J,n}(x)$ obtained when replacing the $X_i$'s by the $X'_i$'s in \eqref{eqn:defEmp}.
The (computationally feasible) estimator $\bar{D}'_{J,n}(x)$ of $\bar{D}_{J}(x,\; P)$ is constructed in a similar manner. The result stated below shows that this approximation stage preserves almost-sure consistency.


\begin{theorem}\label{Approx} Let $J\leq n$. Suppose that, as $n\rightarrow \infty$, 
$$\delta=\max_{1\leq i \leq n}\max_{2\leq k\leq p_i}\left\{ t^{(i)}_{k+1}-t^{(i)}_k \right\}\rightarrow 0.$$   As $n\to \infty$, we have, for any $x\in \mathcal{C}([0,1])$, with probability one, 
$$\left\vert D'_{J,n}(x)-D_{J}(x,\; P)\right\vert \rightarrow 0$$
and 
$$\left\vert \bar{D}'_{J,n}(x)-\bar{D}_{J}(x,\; P) \right\vert \rightarrow 0.$$
\end{theorem}

Refer to the Appendix section for the technical proof.
Given the batch of continuous and piecewise linear curves $X'_1,\; \ldots,\; X'_n$, although the computation cost of the area of their convex hull is of order $O(p \log p)$ with $p=\max_i p_i$, that of the U-statistic $D'_{J,n}(x)$ (and \textit{a fortiori} that of $\bar{D}'_{J,n}(x)$) becomes very expensive as soon as $\binom{n}{J}$ is large. As pointed out in \cite{lopez}, even if the choice $J=2$ for statistics of this type, may lead to a computationally tractable procedure, while offering a reasonable representation of the distribution, varying $J$ permits to capture much more information in general. For this reason, we propose to compute an \textit{incomplete} version of the $U$-statistic $D'_{J,n}(x)$ using a basic Monte-Carlo approximation scheme with $K\geq 1$ replications: rather than averaging over all $\binom{n}{J}$ subsets of $\{1,\;\ldots,\; n\}$ with cardinality $J$ to compute $D'_{J,n}(x)$, one averages over $K\geq 1$ subsets drawn with replacement, forming an \textit{incomplete $U$-statistic}, see \cite{Enqvist78}. The same approximation procedure can be applied (in a randomized manner) to each of the $U$-statistics involved in the average $\bar{D}'_{J,n}(x)$, as described  in the Supplementary Material.

\section{Numerical Experiments}\label{section4}

From a practical perspective, this section explores certain properties of the functional depth proposed using simulated data. It also describes its performance compared with the state-of-the-art methods on (real) benchmark datasets. As a first go, we focus on the impact of the choice of the tuning parameter $K$, which rules the trade-off between approximation accuracy and computational burden and parameter $J$. Precisely, it is investigated through the stability of the ranking induced by the corresponding depths. We next investigate the robustness of the ACH depth (ACHD in its abbreviated form), together with its ability to detect abnormal observations of various types. Finally, the ACH depth is benchmarked against alternative depths standing as natural competitors in the functional setup using real datasets.  A simulation-based study of the variance of the ACH depth is postponed to the Supplementary Material.

 \begin{figure}[!h]
\begin{center}
\includegraphics[height=.17\textheight, trim=1.5cm .0cm 0cm 0cm,clip=true]{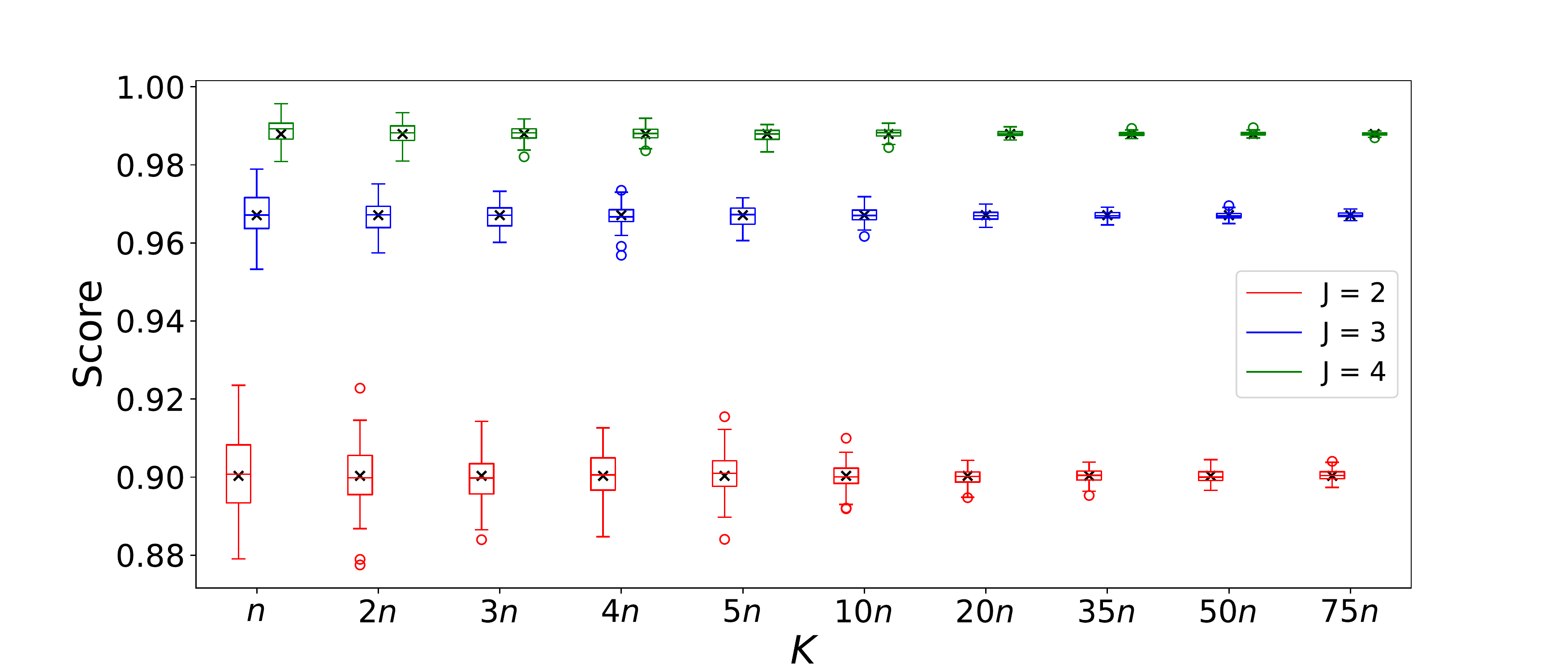}\\\includegraphics[height=.17\textheight, trim=1.5cm .0cm 0cm 0cm,clip=true]{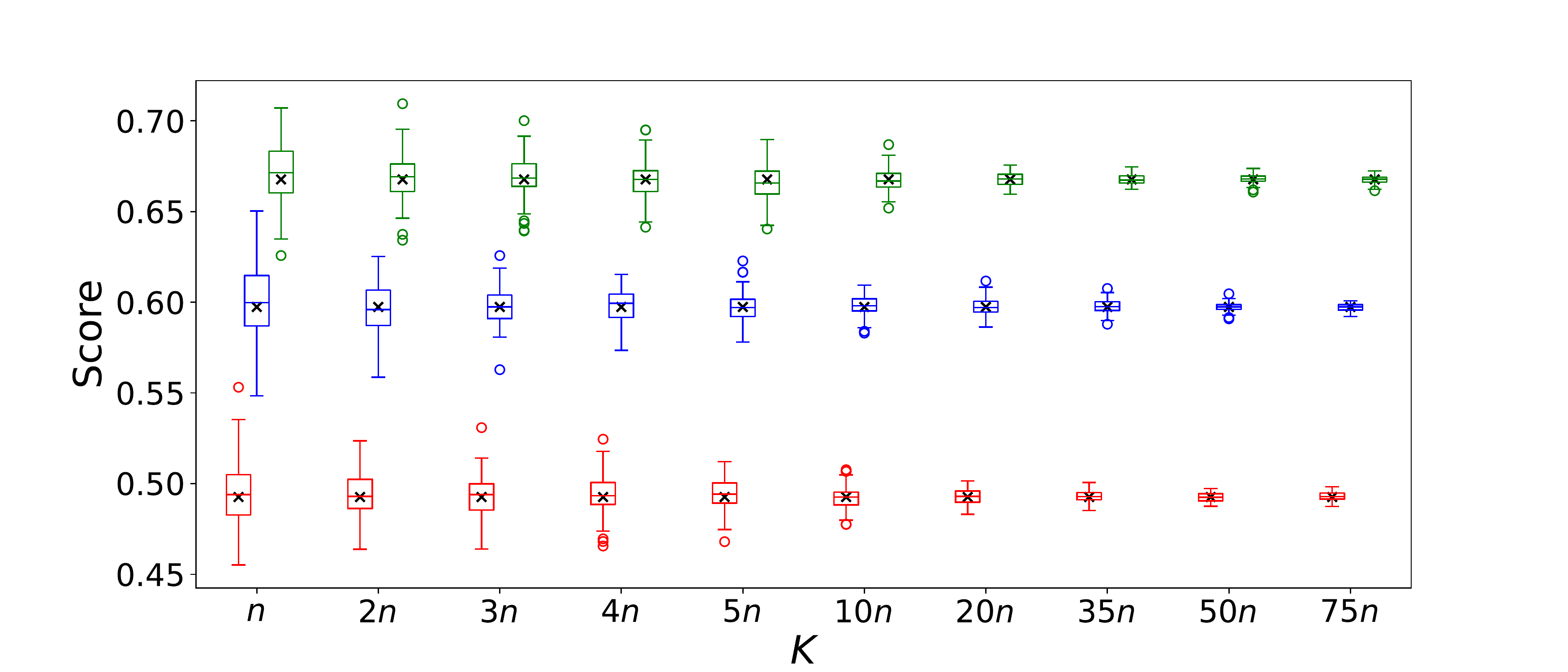}
\end{center}
\caption{Boxplots of the approximations of $D_{J,n} (x_0)$ (top) and $D_{J,n} (x_3)$ (bottom) over different size of $K$. The black crosses correspond to the exact depth measure $D_{J,n}$ for each $J$ respectively.}
\label{fig:impact_K}
\end{figure}

\begin{figure*}[!h]
\begin{center}
\includegraphics[height=.18\textheight, trim=.0cm .0cm 0cm 0cm,clip=true]{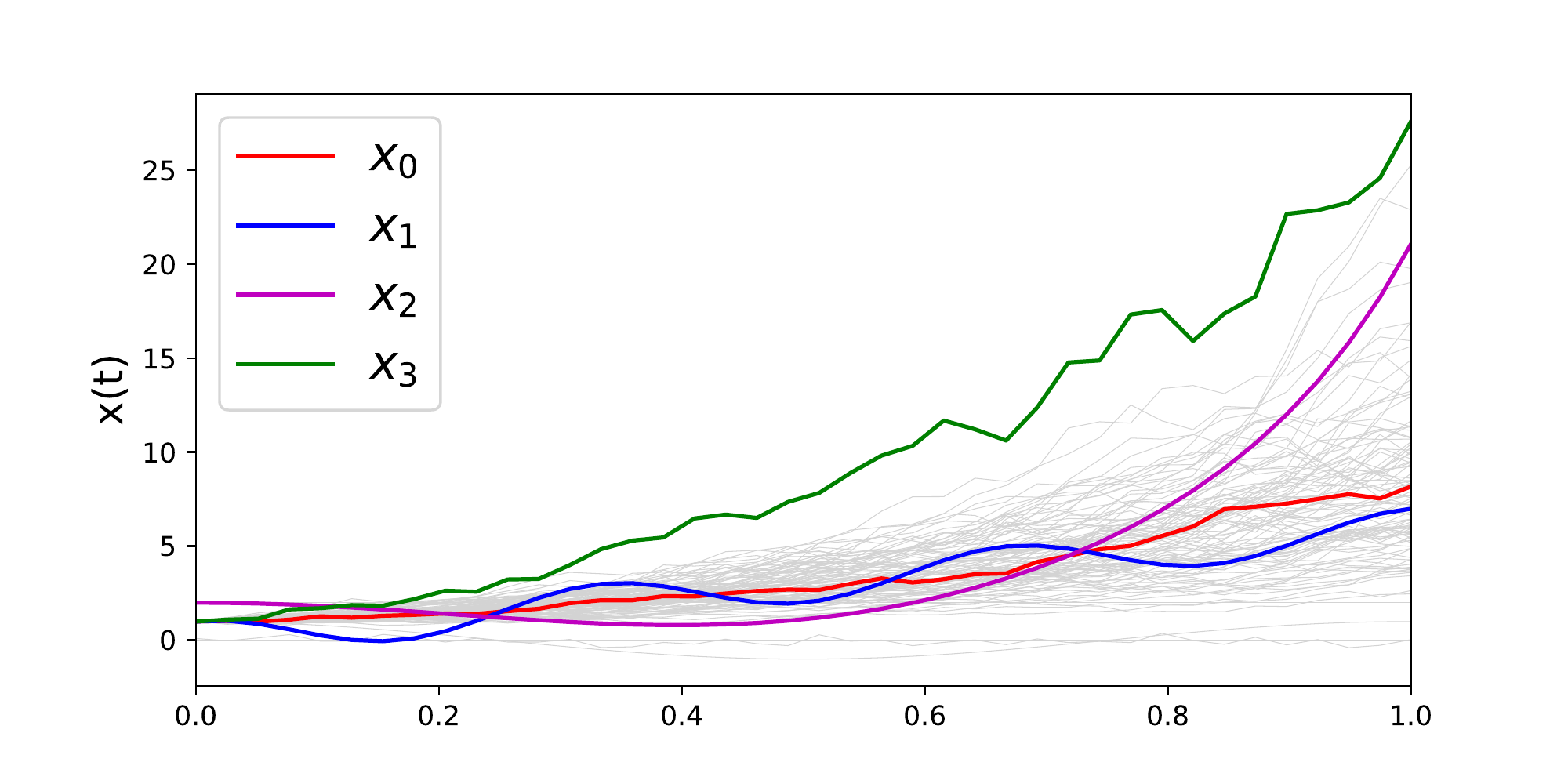}\,\includegraphics[height=.18\textheight, trim=0cm .0cm 0cm 0cm,clip=true]{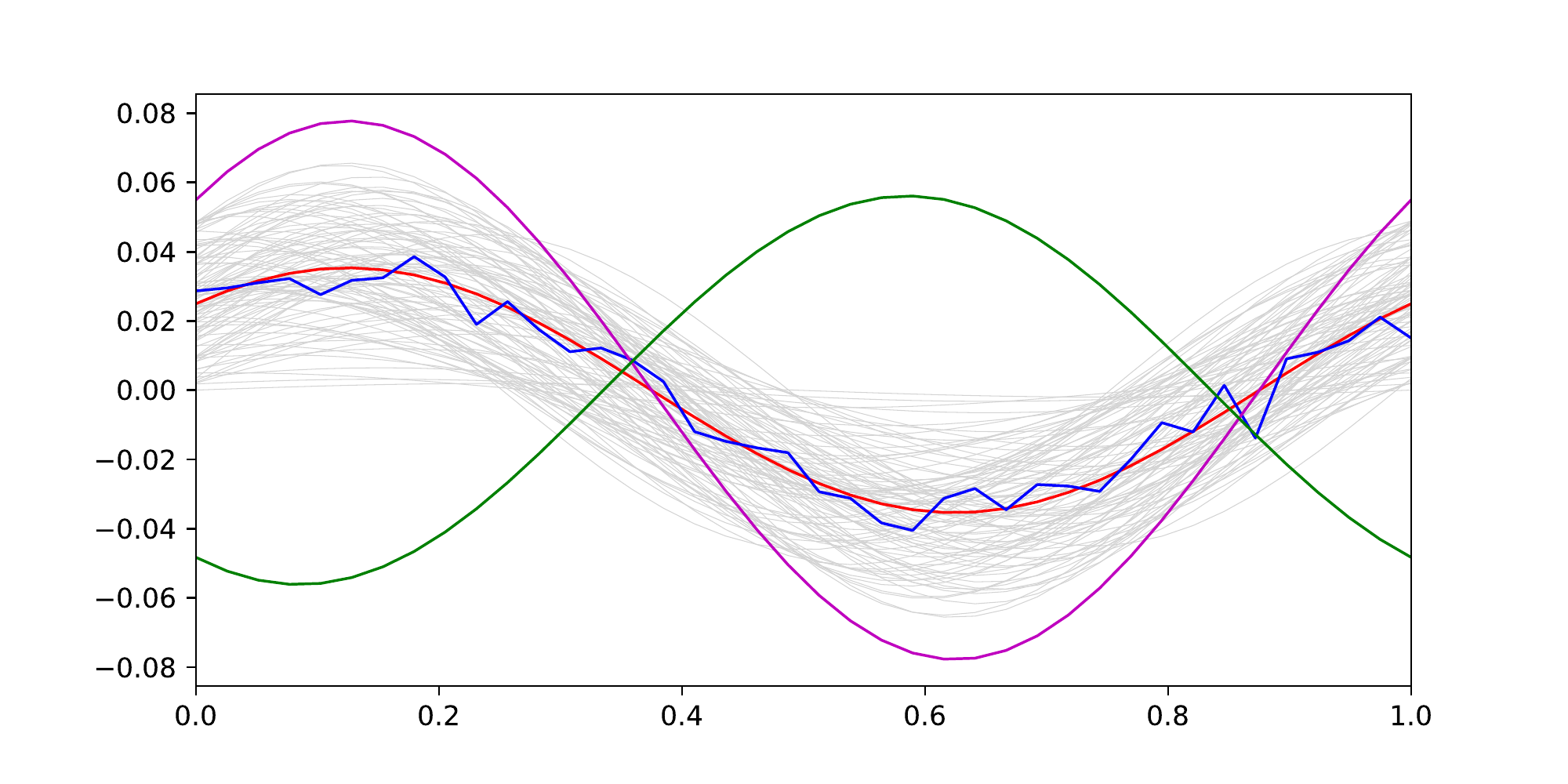}
\end{center}
\caption{Datasets (a) (left) and (b) (right) containing 100 paths with four selected observations. The colors are the same for the four selected observations of both datasets (a) and (b). }
\label{fig:dataset}
\end{figure*}

For the sake of simplicity, the two same simulated datasets, represented in Figure ~\ref{fig:dataset}, are used throughout the section. The dataset $(a)$ corresponds to sample path segments of the \emph{geometric Brownian motion} with mean $2$ and variance $0.5$, a stochastic process widely used in statistical modeling. The dataset $(b)$ consists of smooth curves given by $x(t)=a \cos (2\pi t)+b \sin (2\pi t)$, $t\in[0,1]$, where $a$ and $b$ are independently and uniformly distributed on $[0,0.05]$, as proposed by \cite{claeskens}. Four curves $\{x_i:\; i\in\{0,1,2,3\}\}$ have been incorporated to each dataset: a deep curve and three atypical curves (\textit{anomalies}), with expected depth-induced ranking $D_{J}(x_3) <  D_{J}(x_2)\approx D_{J}(x_1)<D_{J}(x_0)$.

\subsection{Choosing Tuning Parameters $K$ and $J$}\label{subsec:AS}

\begin{figure}[!b]
\includegraphics[trim=1.5cm 0 2.5cm 0,clip=true,scale=0.315]{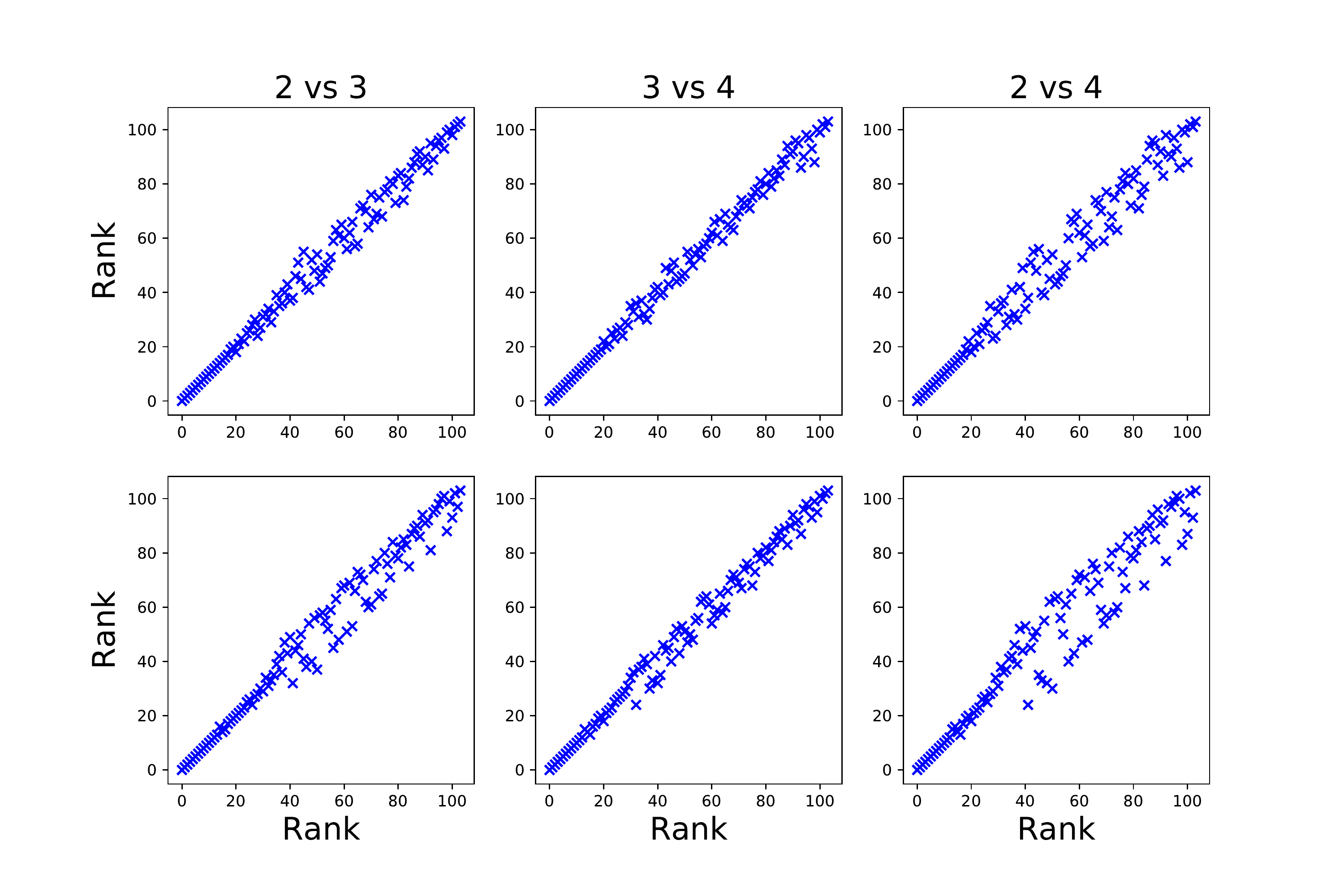}
\caption{Rank-Rank plot for different values of J (2, 3 and 4). The first line represents the rank  over the dataset (a) while the second line represents the dataset (b).}
\label{rankrank}
\end{figure}

Parameter $K$ reflects the trade-off between statistical performance and computational time. In order to investigate its impact on the stability of the method, we compute depths of the deepest and most atypical curves ($x_0$ and $x_3$) for dataset (b), taking $J=2,3,4$. Figure~\ref{fig:impact_K} presents boxplots of the approximated ACHD (together with the exact values of ACHD) over $100$ repetitions. Note that, as expected, depth values grow with $J$. The variance of the depth decreases taking sufficiently small values for $K=5n$ and almost disappearing for $K\ge20n$, while decreasing pattern remains the same for different values of $K$. For these reasons, we keep $K=5n$ in what follows.


The choice of $J$ is less obvious, and clearly when describing an observation in a functional space a substantial part of information is lost anyway. Nevertheless, one observes that computational burden increases exponentially with $J$ and thus smaller values are preferable. Figure~\ref{rankrank} shows the rank-rank plots of datasets (a) and (b) for small values of $J=2,3,4$ and indicates, that depth-induced ranking does not change much with $J$. Thus, for saving computational time, we use value $J=2$ in all subsequent experiments.

\subsection{Robustness}\label{subsec:Rob}

{\renewcommand{\arraystretch}{1} 
\begin{table*}[h]
\begin{tabular}{ | p{1.8cm}| p{2cm}|| p{0.2cm}|p{1.9cm}|  p{1.9cm}| p{1.9cm}| p{1.9cm}| p{1.9cm}|}
 \hline
 \multicolumn{8}{|c|}{$d_{\tau}(\sigma_0, \sigma_{\alpha})( \times 10^{-2}) $} \\
 \hline
& $\boldsymbol{ \alpha }$ & 0& 5 & 10 & 15& 25 & 30 \\
 \hline
 \hline
\multirow{3}{*}{ACHD} &Location& 0 & \bf{0.6} & \bf{1.3} & \bf{2.2} & \bf{4.3} & {\bf 5.2}\\
& Isolated & 0 & {\bf 0.3} & {\bf 1.3} & {\bf 0.9} & {\bf 1.6} &{\bf  2.4} \\
&Shape   &0 & { \bf 0.9} & { \bf 2} & { \bf 2.6} & { \bf 4.2}  &{ \bf  4.7} \\
\hline
\multirow{3}{*}{FSDO} &Location& 0   &3.6 & 7.3 & 10 & 16 & 20 \\
& Isolated &0   & 0.8 & 3.6 & 3.2 & 7.2 & 9.4 \\
&Shape   &0   & 1.6 & 2.9 & 4.2 & 6.6 & 7.4\\
 \hline
\multirow{3}{*}{FT} &Location&0 &5.1 & 9.5 &13& 20 &23\\
& Isolated & 0 &0.7 & 2.7 & 2.7 & 5.9 & 7.2 \\
&Shape   & 0   & 1.7 & 2.9 & 4.3 & 6.6 & 7.7  \\
 \hline
\multirow{3}{*}{FIF} &Location&0  & 7 & 8.2 & 7.3 & 7.3 & 8.9 \\
& Isolated & 0 &  9.3 & 12 & 11 & 10 & 12  \\
&Shape   &  0 &  7.4 & 7.9 & 10 &14 & 14\\
 \hline
\end{tabular}
\caption{Kendall's tau distances between the rank returned with normal data ($\sigma_0$) and contamined data ($\sigma_{\alpha}$, over different portion of contamination $\alpha$ with location, isolated and shape anomalies) for ACHD and three state-of-the-art methods. Bold numbers indicate best stability of the rank over the contaminated datasets.}
\label{rank}
\end{table*}
}

Under robustness of a statistical estimator on understands its ability not to be ``disturbed'' by atypical observations. We explore robustness of ACHD in the following simulation study: between the original dataset and the same dataset contaminated with anomalies, we measure (averaged over $10$ random repetitions) Kendall's $\tau$ distance of two depth-induced rankings $\sigma$ and $\sigma^\prime$, respectively, of the original data:

$$d_{\tau}(\sigma, \sigma ')=\frac{n(n-1)}{2} \sum_{i<j} \mathbbm{1}_{\{(\sigma (i)-\sigma (j))(\sigma '(i)-\sigma '(j))<0 \}}\,.$$

In their overview work, \cite{rousseeuw} introduce taxonomy for atypical observations, focusing on \emph{Location}, \emph{Isolated}, and \emph{Shape} anomalies. Here, we add \emph{Location} anomalies to dataset (a) and \emph{Isolated} and \emph{Shape} anomalies to dataset (b); other types of anomalies for both datasets can be found in the Supplementary Material. The abnormal functions are constructed as follows. \emph{Location} anomalies for dataset (a) are $\tilde{x}(t) = x(t) +a x(t)$ with $a$ drawn uniformly on $[0,1]$. \emph{Isolated} anomalies for dataset (b) are constructed by adding a peak at $t_0$ (drawn uniformly on $[0,1]$) of amplitude $b$ (drawn uniformly on $[0.03,0.06]$) such that $\tilde{y}(t_0)=y(t_0)+b $ and $\tilde{y}(t)=y(t)$ for any $t\neq t_0$. \emph{Shape} anomalies for dataset (b) are $\tilde{z}(t)=z(t)+ 0.01 \times \cos (2\pi t f)+0.01 \times \sin (2\pi t f)$ with $f$ drawn uniformly from $\{1,2,...,10\}$. By varying the percentage of abnormal observations $\alpha$, we compare ACHD to several of the most know in the literature depth approaches: the \emph{functional Stahel-Donoho depth} (FSDO) \citep{rousseeuw} and the \emph{functional Tukey depth} (FT) \citep{claeskens}, and also to the \emph{functional isolation forest} (FIF) algorithm \citep{FIF} which proves satisfactory anomaly detection; see Table~\ref{rank}. One can observe that ACHD consistently preserves depth-induced ranking despite inserted abnormal observation, even if their fraction $\alpha$ reaches $30\%$. FSDO behaves competitively giving slightly better results than ACHD for shape anomalies.

\subsection{Applications to Anomaly Detection}\label{subsec:AD}

\begin{figure}[h]
\includegraphics[trim=3cm 0 0 0,clip=true,scale=0.235]{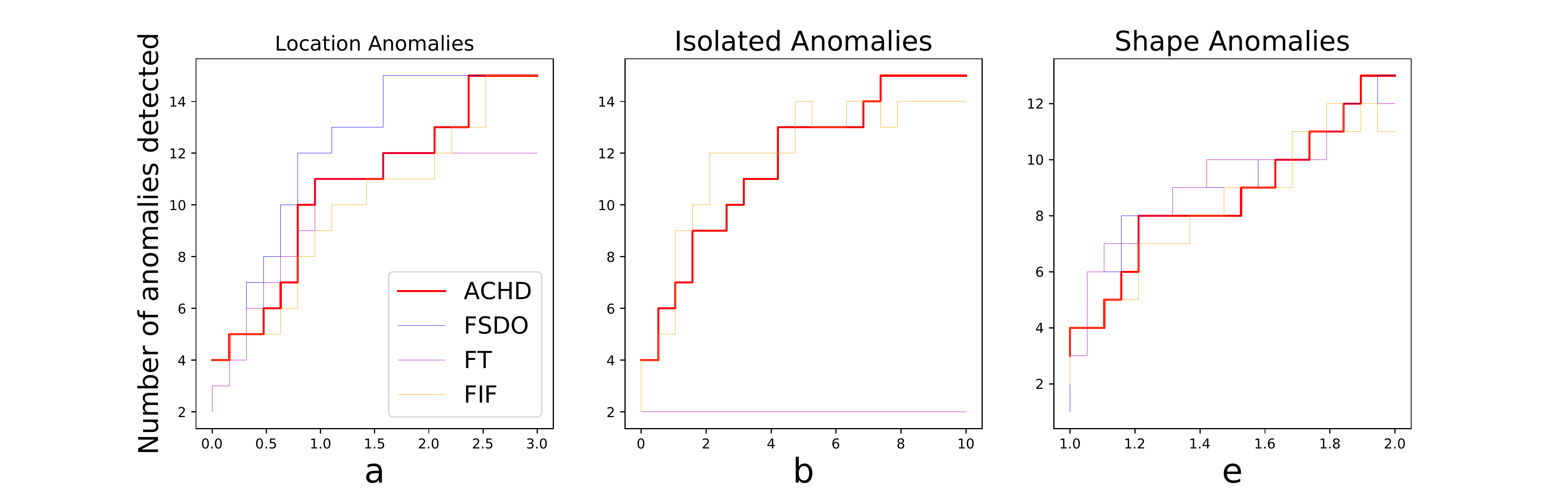}
\caption{Number of anomalies detected over a grid of parameters for three types of anomalies (location, isolated, and shape) for ACHD and three further state-of-the-art methods. }
\label{anom}
\end{figure}
Further, we explore the ability of ACHD to detect atypical observations. For this, we conduct an experiment in settings similar to those in Section~\ref{subsec:Rob}, while changing degree of abnormality gradually for $15$ (out of $100$ curves) in dataset (a). Thus, we alter $a$ in $[0,3]$ for \emph{location anomalies}, $b$ in $[0,10]$ for \emph{isolated anomalies}, and $e$ in $[1,2]$ for \emph{shape anomalies} to amplify the "spikes" of oscillations such that $\tilde{z}(t) =  e z(t)$. (For an illustration of abnormal curves the reader is referred to the Supplementary Material.) Figure~\ref{anom} illustrates number of anomalies detected by ACHD, FSDO, FT, and FIF for different parameters of abnormality. While it is difficult to find the general winner, ACHD behaves favorably in all the considered cases and clearly outperforms the two other depths when the data is contaminated with \emph{isolated anomalies}.



We conclude this section with a \emph{real-world data benchmark} based on three datasets: Octane \citep{Ebensen06}, Wine \citep{Larsen06}, and EOG \citep{UCRArchive}. The Wine dataset consists of 397 measurements of proton nuclear magnetic resonance (NMR) spectra of 40 different wine samples, the Octane dataset are 39 near infrared (NIR) spectra of gasoline samples with 226 measurements, while the EOG dataset represents the electrical potential between electrodes placed at points close to the eyes with 1250 measurements. (Graphs of the three datasets can be found in the Supplementary Material.) As pointed out by \cite{rousseeuw}, it is difficult to detect anomalies in the first two datasets, while they are easily seen during the human eye inspection. For the EOG dataset, we assign smaller of the two classes to be abnormal. To the existing state-of-the-art methods, we add here \emph{Isolation Forest} (IF) \citep{LiuTZ08} and the \emph{One-Class SVM} (OC) \citep{SPSSW01}---multivariate methods applied after a proper dimension reduction (to the dimension 10) using \emph{Functional Principal Component Analysis} (FPCA).\citep{ramsey2}. Portions of detected anomalies (by all the considered methods), indicated in Table~\ref{anomreal}, hint on very competitive performance of ACHD in the addressed benchmark.

\begin{table}[h] 
{\renewcommand{\arraystretch}{1} 
{\setlength{\tabcolsep}{0.18cm}
\begin{tabular}{|c||c|c|c|c|c|c|}
\hline
&ACHD&FSDO&FT&FIF&IF & OC \\
\hline
Octane & {\bf 1} & 0.5 & 0.33 & {\bf 1} & 0.5 & 0.5  \\
\hline
Wine & { \bf 1} & 0 & 0 & { \bf 1}  & 0 & { \bf 1}   \\
\hline
EOG &  { \bf 0.73} & 0.55 & 0.48 & 0.43 & 0.63 & 0.6 \\
\hline
\end{tabular}
}}
\caption{Portion of detected anomalies of benchmark methods for the Octane, Wine, and EOG datasets.}
\label{anomreal}
\end{table}

\section{Conclusion}\label{section5}

In this paper, we have introduced a novel functional depth function on the space $\mathcal{C}([0,1])$ of real valued continuous curves on $[0,1]$ that presents various advantages. Regarding interpretability first, the depth computed at a query curve $x$ in $\mathcal{C}([0,1])$ takes the form of an expected ratio, quantifying the relative increase of the area of the convex hull of i.i.d. random curves when adding $x$ to the batch. We have shown that this depth satisfies several desirable properties and have explained how to solve approximation issues, concerning the sampled character of observations in practice and scalability namely. Numerical experiments on both synthetic and real data have highlighted a number of crucial benefits: reduced variance of its statistical versions, robustness with respect to the choice of tuning parameters and to the presence of outliers in the training sample, capacity of detecting (possibly slight) anomalies of various types, surpassing competitors such as depths of integral type, for isolated anomalies in particular. The open-source implementation of the method, along with all reproducing scripts, can be accessed at \url{https://github.com/GuillaumeStaermanML/ACHD}.

\vspace*{0.2cm}

\noindent {\large \bf Acknowlegments}

The authors thank Stanislav Nagy for his helpful remarks. The authors thank Stanislav Nagy for some helpful remarks. This work has been funded by BPI France in the context of the PSPC Project Expresso (2017-2021).

\bibliographystyle{apalike}
\bibliography{Robust_FSD_arxiv}

\begin{thebibliography}{}

\bibitem[Arcones, 2003]{arcones}
Arcones, M. (2003).
\newblock On the asymptotic accuracy of the bootstrap under arbitrary
  resampling size.
\newblock {\em Ann. Inst. Statist. Math.}, 55:563--583.

\bibitem[Chakraborty and Chaudhuri, 2014]{chak2}
Chakraborty and Chaudhuri (2014).
\newblock The spatial distribution in infinite dimensional spaces and related
  quantiles and depths.
\newblock {\em The annals of statistics}.

\bibitem[Chen et~al., 2015]{UCRArchive}
Chen, Y., Keogh, E., Hu, B., Begum, N., Bagnall, A., Mueen, A., and Batista, G.
  (2015).
\newblock The ucr time series classification archive.

\bibitem[Claeskens et~al., 2014]{claeskens}
Claeskens, G., Hubert, M., Slaets, L., and Vakili, K. (2014).
\newblock Multivariate functional halfspace depth.
\newblock {\em Journal of American Statistical Association}, 109(505):411--423.

\bibitem[Cuevas et~al., 2007]{cuevas}
Cuevas, A., Febrero, M., and Fraiman, R. (2007).
\newblock Robust estimation and classification for functional data via
  projection-based depth notions.
\newblock {\em Computational Statistics}, 22(3):481--496.

\bibitem[Dudley, 1984]{dudley2}
Dudley (1984).
\newblock A course on empirical processes.
\newblock In springer, editor, {\em Ecole d'Eté de probabilités de
  Saint-Flour XII. Lecture Notes in Math}. Cambridge University press, New
  York.

\bibitem[Dudley, 2002]{dudley}
Dudley, R.~M. (2002).
\newblock {\em Real Analysis and Probability}.
\newblock Cambridge Studies in Advanced Mathematics. Cambridge University
  Press, 2 edition.

\bibitem[Dutta et~al., 2011]{dutta}
Dutta, Ghosh, and Chaudhuri (2011).
\newblock Some intriguing properties of tukey’s half-space depth.
\newblock {\em Bernouilli}.

\bibitem[Enqvist, 1978]{Enqvist78}
Enqvist, E. (1978).
\newblock {\em On sampling from sets of random variables with application to
  incomplete {$U$}-statistics}.
\newblock PhD thesis, Lund University.

\bibitem[Esbensen, 2001]{Ebensen06}
Esbensen, K. (2001).
\newblock Multivariate data analysis-in practice.
\newblock {\em Camo Software}.

\bibitem[Ferraty and Vieu, 2006]{ferraty}
Ferraty, F. and Vieu, P. (2006).
\newblock {\em Nonparametric Functional Data Analysis}.
\newblock Springer-Verlag, New York.

\bibitem[Fraiman and Muniz, 2001]{Fraiman2001}
Fraiman, R. and Muniz, G. (2001).
\newblock Trimmed means for functional data.
\newblock {\em Test}, 10(2):419--440.

\bibitem[Gijbels and Nagy, 2018]{Gijbels}
Gijbels, I. and Nagy, S. (2018).
\newblock On a general definition of depth for functional data.
\newblock {\em Statistical Science}.

\bibitem[Hubert et~al., 2015]{rousseeuw}
Hubert, M., Rousseeuw, P.~J., and Segaert, P. (2015).
\newblock Multivariate functional outlier detection.
\newblock {\em Statistical Methods {\&} Applications}, 24(2):177--202.

\bibitem[Jörnsten, 2004]{jornsten}
Jörnsten, R. (2004).
\newblock Clustering and classification based on the l1 data depth.
\newblock {\em Journal of Multivariate Analysis}, 90(1):67 -- 89.

\bibitem[Lange et~al., 2014]{LangeMM14}
Lange, T., Mosler, K., and Mozharovskyi, P. (2014).
\newblock Fast nonparametric classification based on data depth.
\newblock {\em Statistical Papers}, 55(1):49--69.

\bibitem[Larsen et~al., 2006]{Larsen06}
Larsen, F., Berg, F., and Engelsen, S. (2006).
\newblock An exploratory chemometric study of 1 h nmr spectra of table wine.
\newblock {\em Journal of Chemometrics}, 20:198 -- 208.

\bibitem[Lee, 1990]{Lee90}
Lee, A.~J. (1990).
\newblock {\em ${U}$-statistics: Theory and practice}.
\newblock Marcel Dekker, Inc., New York.

\bibitem[Liu et~al., 2008]{LiuTZ08}
Liu, F.~T., Ting, K.~M., and Zhou, Z. (2008).
\newblock Isolation forest.
\newblock In {\em 2008 Eighth IEEE International Conference on Data Mining},
  pages 413--422.

\bibitem[Long and Huang, 2016]{long5}
Long and Huang (2016).
\newblock A study of functional depths.
\newblock {\em preprint}.

\bibitem[Lopez-Pintado and Romo, 2009]{lopez}
Lopez-Pintado, S. and Romo, J. (2009).
\newblock On the concept of depth for functional data.
\newblock {\em Journal of the American Statistical Association}.

\bibitem[Lopez-Pintado and Romo, 2011]{lopez2}
Lopez-Pintado, S. and Romo, J. (2011).
\newblock A half-region depth for functional data.
\newblock {\em Computational Statistics and Data Analysis}.

\bibitem[Mosler and Mozharovskyi, 2017]{mozha}
Mosler, K. and Mozharovskyi, P. (2017).
\newblock Fast dd-classification of functional data.
\newblock {\em Statistical Papers}, 58(4):1055--1089.

\bibitem[Nagy et~al., 2016]{stanislas_approx}
Nagy, S., Gijbels, I., and Hlubinka, D. (2016).
\newblock Weak convergence of discretely observed functional data with
  applications.
\newblock {\em Journal of Multivariate Analysis}, 146:46 -- 62.

\bibitem[Nagy et~al., 2017]{nagy}
Nagy, S., Gijbels, I., and Hlubinka, D. (2017).
\newblock Depth-based recognition of shape outlying functions.
\newblock {\em Journal of Computational and Graphical Statistics},
  26(4):883--893.

\bibitem[Nieto-Reyes and Battey, 2016]{nieto}
Nieto-Reyes, A. and Battey, H. (2016).
\newblock A topologically valid definition of depth for functional data.
\newblock {\em Statistical Science}.

\bibitem[Oja, 1983]{oja}
Oja (1983).
\newblock Descriptive statistics for multivariate distributions.
\newblock {\em Statistics and Probability Letters}.

\bibitem[Ramsay and Silverman, 2002]{ramsey2}
Ramsay, J.~O. and Silverman, B.~W. (2002).
\newblock {\em Applied Functional Data Analysis: Methods and Case Studies}.
\newblock Springer-Verlag, New York.

\bibitem[Ramsay and Silverman, 2005]{ramsey}
Ramsay, J.~O. and Silverman, B.~W. (2005).
\newblock {\em Functional Data Analysis}.
\newblock Springer-Verlag, New York.

\bibitem[Rossi and Villa, 2006]{SVMF}
Rossi, F. and Villa, N. (2006).
\newblock Support vector machine for functional data classification.
\newblock {\em Neurocomputing}, 69(7):730 -- 742.

\bibitem[Schneider, 2013]{schneider2}
Schneider, R. (2013).
\newblock {\em Convex Bodies: The Brunn-Minkowski Theory}.
\newblock Cambridge University Press, Cambridge.

\bibitem[Schneider and Weil, 2008]{schneider}
Schneider, R. and Weil, W. (2008).
\newblock {\em Stochastic and Integral Geometry}.
\newblock Springer-Verlag, Berlin Heidelberg.

\bibitem[Sch\"olkopf et~al., 2001]{SPSSW01}
Sch\"olkopf, B., Platt, J., Shawe-Taylor, J., Smola, A., and Williamson, R.
  (2001).
\newblock Estimating the support of a high-dimensional distribution.
\newblock {\em Neural Computation}, 13(7):1443--1471.

\bibitem[Serfling, 2006]{ser2006}
Serfling, R. (2006).
\newblock Depth functions in nonparametric multivariate inference.
\newblock {\em DIMACS Series in Discrete Mathematics and Theoretical Computer
  Science}, 72.

\bibitem[Staerman et~al., 2019]{FIF}
Staerman, G., Mozharovskyi, P., Cl\'{e}mençon, S., and d'Alch\'{e} Buc, F.
  (2019).
\newblock Functional isolation forest.
\newblock In {\em Proceedings of The 11th Asian Conference on Machine
  Learning}.

\bibitem[Tukey, 1975a]{tukey}
Tukey (1975a).
\newblock Mathematics and the picturing of data.
\newblock {\em In Proceedings of the International Congress of Mathematicians}.

\bibitem[Tukey, 1975b]{Tukey75}
Tukey, J.~W. (1975b).
\newblock Mathematics and the picturing of data.
\newblock In James, R., editor, {\em Proceedings of the International Congress
  of Mathematicians}, volume~2, pages 523--531. Canadian Mathematical Congress.

\bibitem[Zuo and Serfling, 2000]{ZuoSerfling00}
Zuo, B. and Serfling, R. (2000).
\newblock General notions of statistical depth function.
\newblock {\em The Annals of Statistics}, 28(2):461--482.

\end{thebibliography}

\clearpage

\setcounter{section}{0}
\renewcommand{\thesection}{\Alph{section}}

\onecolumn

\aistatstitle{ \Large Supplementary Material to the Article \\ The Area of the Convex Hull of Sampled Curves:\\
a Robust Functional Statistical Depth Measure}

First, Section~\ref{sec:proofs} collects technical proofs omitted in the body of the article. Then, Section~\ref{sec:algoappr} provides the exact algorithm for approximate computation of the proposed depth notion. Finally, Section~\ref{sec:addexper} collects further experimental results mentioned in the article.

\section{Technical proofs}\label{sec:proofs}
This part presents the proofs of Proposition~3.1, Theorems~3.1 and~3.2 as well as the counter examples for the non-satisfied properties. Most of the proofs are done for both  $D_J$ and $\overline{D}_J$.
 
\subsection{Proof of Proposition 3.1}
\subsubsection{Affine-invariance}

Let $a,b\in \mathbb{R}$, it is clear that 
\begin{align*}
&conv\left(graph \left(\{aX_1+b,...,aX_n+b \} \right) \right)\\&= a\times conv \left(graph\left(\{X_1,...,X_n \} \right) \right)+b
\end{align*} where $ a \times    conv\left( graph\left(\{X_1,...,X_n \} \right) \right)+b=\lbrace (t,ax+b): (t,x)\in  conv \left(graph\left(\{X_1,...,X_n \} \right) \right) \rbrace$. Following this, and by properties of Lebesgue measure, we have 
\begin{align*}
\frac{\lambda_2\left(a \times conv \left(graph \left(\{X_1,...,X_n \} \right) \right)+b \right)}{\lambda_2\left(a \times conv  \left(graph \left(\{X_1,...,X_n,z \} \right) \right)+b \right)}&=\frac{\lambda_2\left(a \times conv \left(graph \left(\{X_1,...,X_n \} \right) \right) \right)}{\lambda_2\left(a\times conv \left(graph\left(\{X_1,...,X_n,z \} \right) \right) \right)}\\&=\frac{a \times \lambda_2\left(conv \left(graph \left(\{X_1,...,X_n \} \right) \right) \right)}{a\times \lambda_2\left(conv \left(graph \left(\{X_1,...,X_n,z \} \right) \right) \right)}\\&=\frac{\lambda_2\left(conv \left(graph\left(\{X_1,...,X_n \} \right) \right) \right)}{\lambda_2\left(conv \left(graph \left(\{X_1,...,X_n,z \} \right) \right) \right)}
\end{align*}

\begin{figure}[!h]
\begin{center}
\includegraphics[scale=0.3]{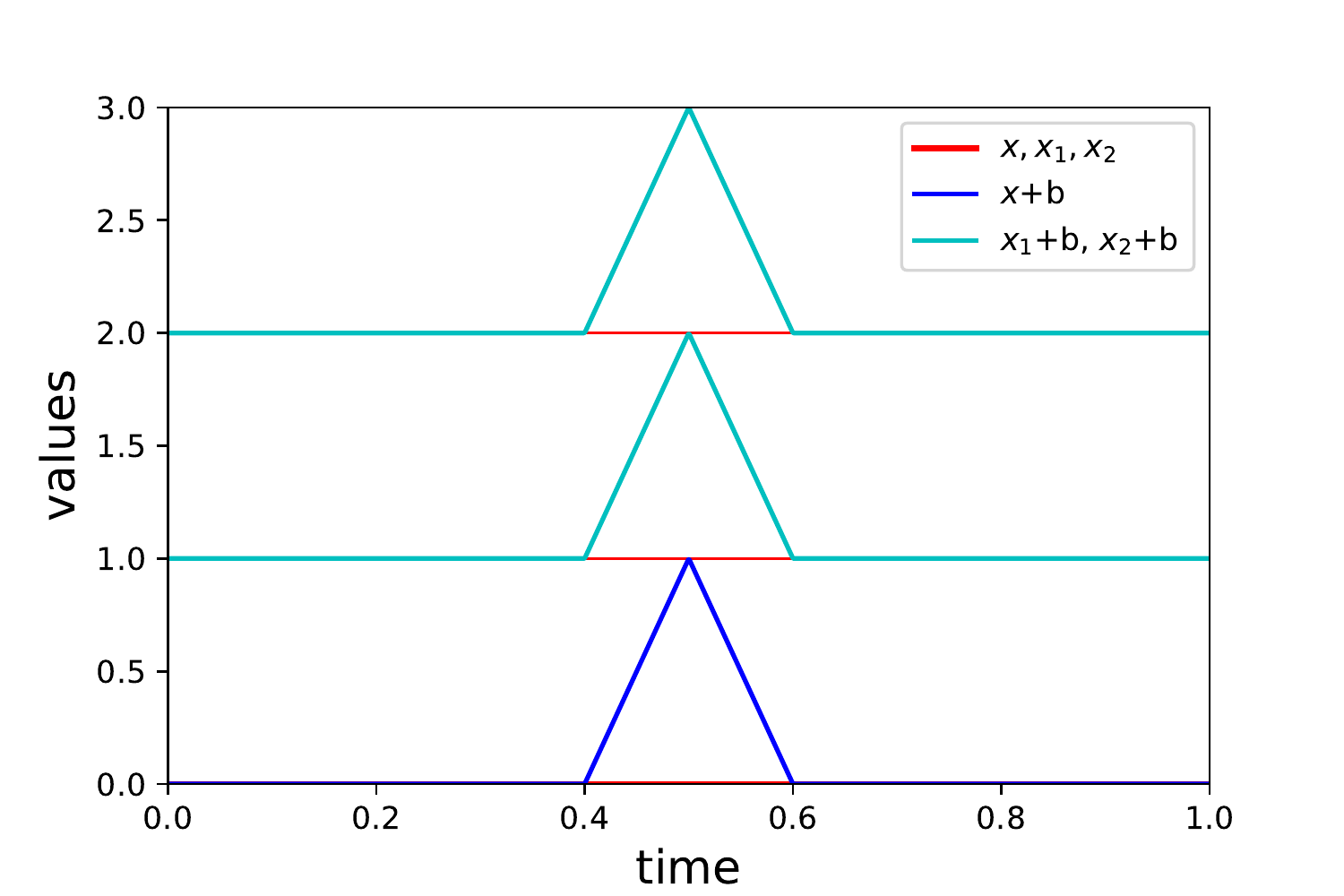}
\end{center}
\caption{Plots of the functions used in the case of $ \mathbf{a},\mathbf{b} \in \mathcal{X}$. The three red lines come from $x_1,x_2,x$. The cyan curves correspond to $x_1+b$ and $x_2+b$ and blue curve to $x+b$.}
\label{fig:count1}
\end{figure}

\noindent \textbf{The case of $ \mathbf{a},\mathbf{b} \in \mathcal{X}$:}\\
Now, we just take a counter example to prove that it is not true if $b$ belongs to $\mathcal{X}$, the case where $a\in \mathcal{X}$ is trivial.
\noindent
For the sake of simplicity, let $I=[0,1]$ and $J=2$. If we take $x\equiv 0, x_1\equiv 1, x_2 \equiv 2$ and $X$ a random variable with distribution $P$ such that $\mathbb{P}(X\equiv x_1)=\frac{1}{2}$ and $\mathbb{P}(X\equiv x_2)=\frac{1}{2}$. Let $X_1,X_2$ be samples from $P$ and $b$ be continuous function $ t \mapsto (10t-4)\mathbbm{1}_{([0.4,0.5])}+(-10t+6)\mathbbm{1}_{([0.5,0.6])}$. It is easy to see that $D_{J}(x|P)=\frac{1}{8}\neq D_{J}(x+b|P_{X+b})$ since 
\begin{align*}
D_{J}(x+b|P_{X+b})&=\frac{1}{2} \times  \left(\frac{1}{2}\times \frac{0.5}{1.5} +\frac{1}{2}\times \frac{0.5}{2.5} \right)^{j=1} +\frac{1}{2} \times \left( \frac{1}{4}\times \frac{0.5}{1.5}+\frac{1}{4}\times \frac{0.5}{2.5}+\frac{1}{2}\times \frac{1.5}{2.5} \right)^{j=2}\\&= \frac{8}{60}+\frac{9}{60}\\&=\frac{17}{60}.
\end{align*}
Note that even if we set $j>1$ to avoid the fact that the convex hull of constant function have null Lebesgue measure, $D_{J}(x|P)$ and $D_{J}(x+b|P_{X+b})$ remain different, see Fig.~\ref{fig:count1}.

\subsubsection{Vanishing at infinity}

Let $J$ be fixed and $x_n$ a sequence of function such that $||x_n||$ tends to infinity when $n$ grows, for every $j \in \{1,...,J \}$ we define :

\begin{align*}
\Phi_{x_n}:  &  \quad\mathcal{X}^{\otimes j} \; \; \; \; \; \; \longrightarrow [0,1] \\&
(x_1,...,x_j)\longmapsto 
 \frac{\lambda_2\left(conv \left(graph (\{x_1,...,x_j \})\right) \right)}{\lambda_2\left(conv \left(graph(\{x_1,...,x_j,x_n \})\right)\right)}
\end{align*}

\noindent
As a continuous function on compact set, $x_1,...,x_j$ are bounded, then $\Phi_{x_n}\underset{||x_n||_{\infty} \rightarrow \infty}{\longrightarrow} 0$. The result follows from dominated convergence theorem since $\Phi_{x_n}$ is bounded by 1.

\subsubsection{Continuity in $x$}\label{P4}
Let $x_1,...,x_j$, $j\in \{1,...,J \}$ be fixed curves in $\mathcal{C}(I)$ with at least two different curves, i.e, there exists a $t\in I$ and $ l,k \in {1,...,j}$ such that $x_k(t)\neq x_l(t)$. If $j=1$, we need that $x_1$ is not a constant function. From Lemma \ref{L}, we know that the function $$f : x \longmapsto Band(x_1,...,x_j,x)$$ is continuous. 

\noindent
Let $\mathcal{K}_2$ be the set of all compact set in $\mathbb{R}^{2}$ and $\mathcal{K}^{C}_2$ the set of all convex bodies (compact, convex set with non-empty interior). We equip both spaces with the Hausdorff distance. We know that : 
\begin{align*}
conv : &\,\mathcal{K}_2 \longrightarrow \mathcal{K}_2  \\&\, K_2 \longmapsto conv(K_2)
\end{align*}
and 
\begin{align*}
\psi: &\,\mathcal{K}^{C}_2 \longrightarrow \mathcal{K}^{C}_2  \\&\, K^{c}_2 \longmapsto \lambda_2(K^{c}_2)
\end{align*}
are continuous with respect to the Hausdorff distance. See for example, Theorems~12.3.5 and~12.3.6 in \cite{schneider} for $g$ and Theorem~1.8.16 in \cite{schneider2} for $\psi$. Then $\Phi:=\psi \circ conv \circ f$ : $x\,\mapsto \lambda_2\left(conv \left(graph \{x_1,...,x_j,x \}) \right)\right) $ is continuous. \\
It is straightforward to show that $$\phi : x \longmapsto \frac{\lambda_2\left(conv \left( graph(\{x_1,...,x_j \}) \right) \right)}{\lambda_2\left(conv \left(graph(\{x_1,...,x_j,x \})\right) \right)} $$ is continuous.
Now, we just have to prove that $$\; x\longmapsto \mathbb{E}\left[  \frac{\lambda_2\left(conv \left( graph(\{x_1,...,x_j \}) \right) \right)}{\lambda_2\left(conv \left(graph(\{x_1,...,x_j,x \})\right) \right)}  \right]$$ is continuous which is true by dominated convergence theorem. We conclude the proof with the continuity of the sum of continuous functions.

\subsection{Continuity in $P$}\label{P6}

 $\left(\mathcal{C}([0,1]),||.||_{\infty} \right)$ is a polish space and implies that the set of all probability measures on this space with the Lévy-Prohorov metric $\rho_{LP}$ is still polish. By the portmanteau theorem (\textit{e.g.}, see Theorem~11.3.3 in \cite{dudley}) it follows that $\rho_{LP}(Q_n,Q)\rightarrow 0$ is equivalent to $Q_n \overset{d}{\rightarrow} Q$ for $Q,Q_n$  respectively a measure and a sequence of measures on $\mathcal{C}([0,1])$. It implies that $$\int fdQ_n \longrightarrow \int fdQ$$ for every $f$ bounded continuous real function on $\left(\mathcal{C}([0,1]),||.||_{\infty} \right)$. \\
 
Let  $j$ be fixed natural number and define the following function 
\begin{align*}
\Phi_x :  & \quad \quad  \mathcal{X}^{\otimes j} \; \;   \longrightarrow [0,1] \\& (x_1,...,x_j)\longmapsto 
 \frac{\lambda_2\left(conv\left(graph(\{x_1,...,x_j \}) \right) \right)}{\lambda_2\left(conv\left(graph(\{x_1,...,x_j,x \}) \right)\right)}
\end{align*}
If we equip $\mathcal{C}([0,1])^{\otimes j}$ with the infinite norm $|||.|||_{\infty,j}$ defined by $$|||f|||_{\infty,j}=max(||f_1||_{\infty},...,||f_j||_{\infty}),$$
following the same argument from the proof \ref{P4}, $\Phi_x$ is bounded and continuous. \\
Now, let $J\leq n$ be fixed and $P_n$ be a sequence of measures on $\mathcal{C}([0,1])$ such that $\rho_{LP}(Q_n,Q)\rightarrow 0$. we have :
\begin{align*}
\underset{n\rightarrow \infty}{\lim}\sum_{j=1}^{J}  D_{j}(x|Q_n)&=\sum_{j=1}^{J} \underset{n\rightarrow \infty}{\lim} \int_{\mathcal{C}([0,1])^{\otimes j}} \Phi_x dQ^{\otimes j}_n\\&= \sum_{j=1}^{J} \int_{\mathcal{C}([0,1])^{\otimes j}} \Phi_x dQ^{\otimes j} \\& = \sum_{j=1}^{J} D_j(x | Q)
\end{align*}
Then the results holds for $\sum_{j=1}^{J} D_j$ (and trivially for $D_J$).

\subsection{Proof of the Theorem 3.1}\label{A1}
\noindent

\noindent

\noindent

For every $1 \leq j \leq J$,the term $ | D_{j,n}(x|P_n)-D_{j}(x|P)  | $ goes to zero almost-surely by U-statistics consistency (see \textit{e.g.} \cite{hoeffding1961strong}).
Then 
$$\mathbb{P}\left(\forall j : \bigg |D_{j,n}(x|P_n)- D_{j}(x|P) \bigg|\rightarrow 0 \right)=1$$

which is equivalent to
$$\mathbb{P}\left(\sum_{j=1}^{J}\bigg |D_{j,n}(x|P_n)- D_{j}(x|P) \bigg |\rightarrow 0 \right)=1\,.$$

By triangle inequality, for any $x\in \mathcal{C}([0,1])$,
 \begin{align*}
\bigg |\sum_{j=1}^{J} D_{j,n}(x|P_n)-D_{j}(x|P) \bigg | &\leq \sum_{j=1}^{J} \bigg | D_{j,n}(x|P_n)-D_{j}(x|P)\bigg | \\& \leq   \sum_{j=1}^{J}  \bigg | D_{j,n}(x|P_n)-D_{j}(x|P) \bigg | 
\end{align*}
which leads to the result.


\subsection{Proof of the Theorem 3.2}

The result follows from the continuity in $P$ and Theorem~3 in \cite{stanislas_approx}.

\subsection{Counter-examples for the not-satisfied properties}
\subsubsection{Maximality at the center}
We restrict ourselves for simplicity to $J=2$ and $I=[0,1]$. Let $X \sim P$ be a distribution such that $P(X \equiv y_1)=P(X \equiv y_2)=\frac{1}{2}$ with 
\begin{align*}
y_1 &= (-2t+1)\mathbbm{1}_{[0,0.25]}+ (2t)\mathbbm{1}_{[0.25,0.5]} + (-2t+2)\mathbbm{1}_{[0.5,0.75]}+ (2t-1)\mathbbm{1}_{[0.75,1]}\,, \\&\hspace*{-0.3cm} y_2=-y_1\,.
\end{align*}

\noindent
The distribution is clearly centrally and halfspace symmetric around $\theta \equiv 0$ but we have 
$$D_{J}(\theta,P)<D_{J}(y_1,P)=D_{J}(y_2,P)\,.$$
\noindent
Since 
\begin{align*}
D_{J}(0,P)&=\frac{1}{2}\times \left(\frac{1}{2} \times \frac{3}{8}+\frac{1}{2} \times \frac{3}{8} \right)^{j=1} +\frac{1}{2}\times \left(\frac{1}{2}+\frac{1}{4} \times \frac{3}{8}+\frac{1}{4} \times \frac{3}{8} \right)^{j=2} \\& =\frac{17}{32}\approx 0.53
\end{align*}
and 
\begin{align*}
D_{J}(y_1,P)&=\frac{1}{2}\times \left(\frac{1}{2} \times +\frac{1}{2} \times \frac{3}{16} \right)^{j=1} +\frac{1}{2}\times \left(\frac{1}{2}+\frac{1}{4} +\frac{1}{4} \times \frac{3}{16} \right)^{j=2} \\& =\frac{70}{128}\approx 0.546
\end{align*}

\begin{figure}[!h]
\begin{center}
\includegraphics[scale=0.3]{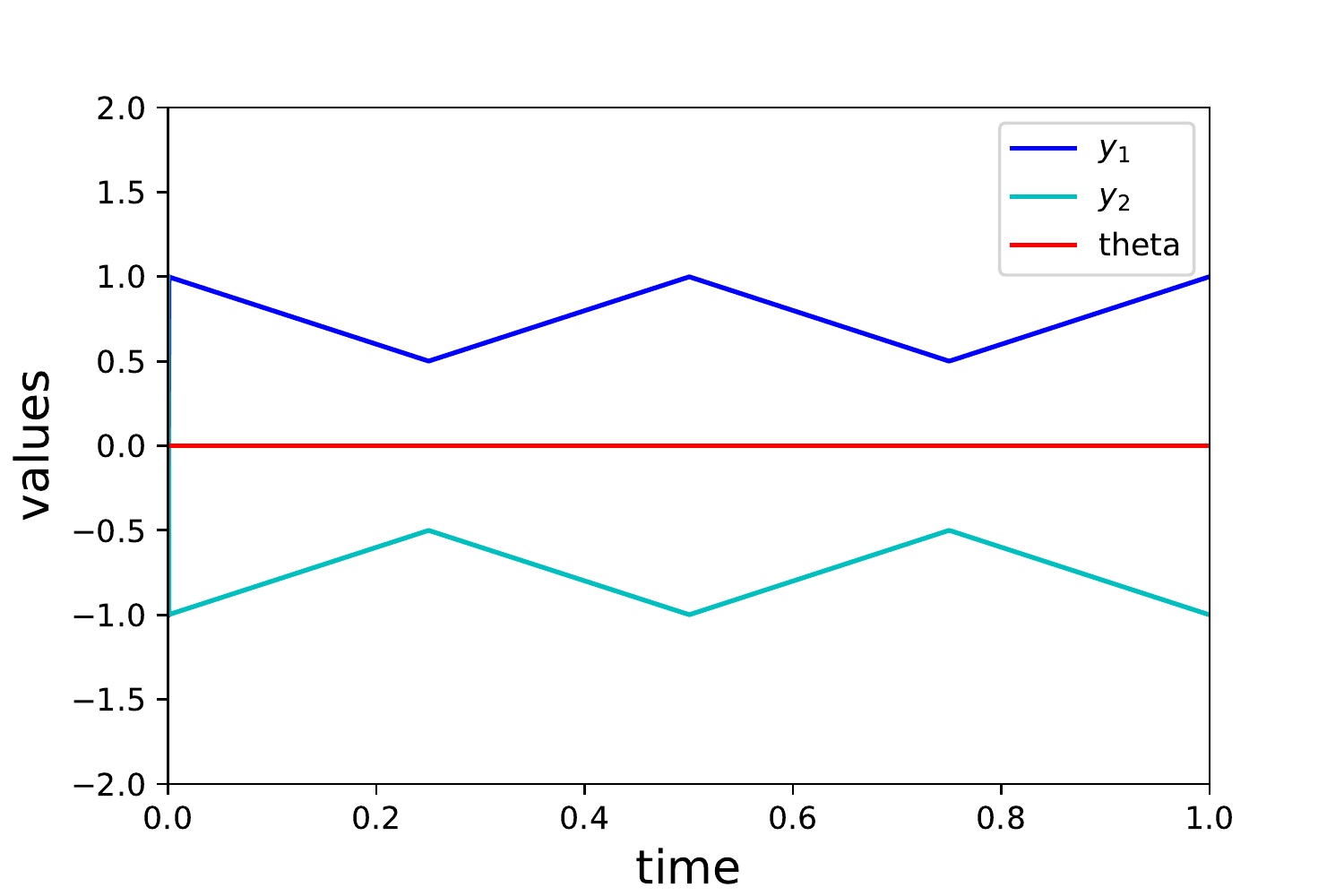}
\end{center}
\caption{Plot of the functions used in the counter example of the maximality at the center property. $y_1$ (blue curve) and $y_2$ (cyan curve) correspond to the distribution and $\theta \equiv 0$ corresponds to the red curve.  }
\end{figure}
\subsubsection{Decreasing w.r.t. the deepest point}
We restrict ourselves $I$ to $[0,1]$ and $J=2$ for the sake of comprehension but the example still works on every $I$.  Let $P$ be the distribution  such that $$P(X \equiv 0)=P(X \equiv 1)=P(X \equiv -1)=\frac{1}{3}.$$ It is clear from this distribution that $0 \in \underset{x \in \mathcal{X}}{\text{sup }}D(x,P)$ and , if we write $z\equiv 0$, $ D_{J}(z,P)=\frac{1}{4}$. We define $y \equiv 1.5$ and $x(t) = 4t\mathbbm{1}_{[0,0.5]}(t) +(-4t+4)\mathbbm{1}_{[0.5,1]}(t)$. We have $d(x,z)=2$, $d(x,y)=0.5$ and $d(y,z)=1.5$. If we compute the depth of $x$ and $y$ we have :
$$D_{J}(y,P)=\frac{1}{2}\times \frac{2}{9}\times\left( \frac{4}{5}+ \frac{2}{5}+\frac{2}{3} \right)=\frac{23}{135}$$
and $$D_{J}(x,P)=\frac{1}{2}\times \frac{2}{9}\times\left( \frac{4}{5}+ \frac{1}{2}+\frac{8}{9} \right)=\frac{197}{810}\,.$$
The result follows. Notice that the result remains true if $conv$ is replaced by the $band$ function.
\begin{figure}[!h]
\begin{center}
\includegraphics[scale=0.3]{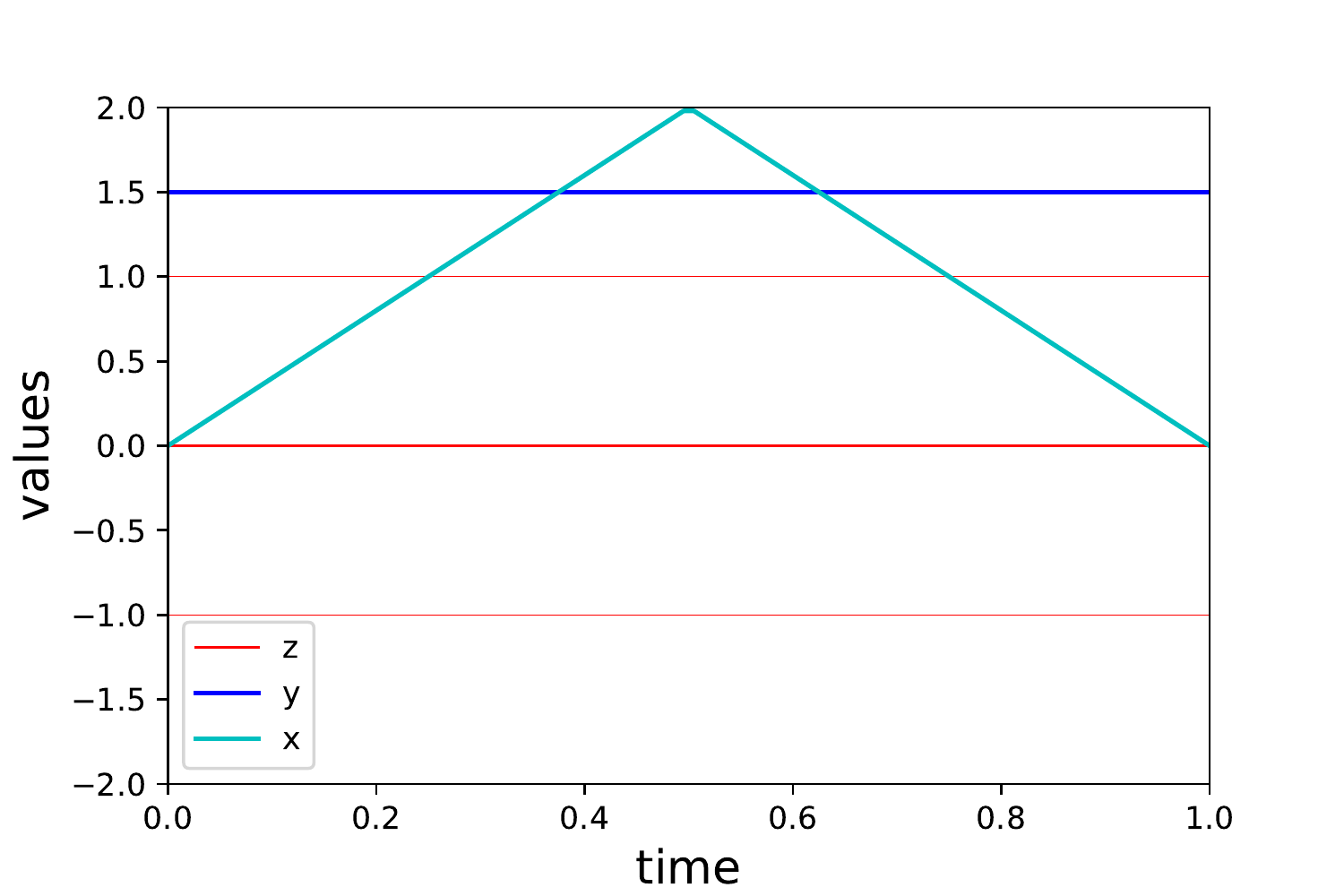}
\end{center}
\caption{Plots of the functions used in the counter example of the decreasing property. The three red lines come from the distribution, the thicker red curve corresponds to the maximal depth. The cyan curve corresponds to $x$ and the blue curve to $y$. }
\end{figure}

\subsection{Technical requirements}

\begin{lemma} \label{L} Let $x_1,...,x_j$, $j\in \{1,...,J \}$ be fixed curves in $\mathcal{C}([0,T])$. The function
\begin{align*} &\mathcal{C}([0,T]) \longrightarrow \mathcal{K}_2
 \\& x \mapsto Band(x_1,...,x_j,x)
\end{align*} 
is continuous if we equip $\mathcal{K}_2$ with the Hausdorff distance $d_H$.
\end{lemma}

\noindent \textbf{Proof.} Let $x_0 \in \mathcal{C}([0,T])$  and $j$ be fixed in $\{1,...,J \}$.
We want to prove $$\forall \epsilon>0, \exists \delta : \forall x \in \mathcal{C}([0,T]) : ||x-x_0||_{\infty}<\delta \Rightarrow d_H(Band(x_1,...,x_j,x),Band(x_1,...,x_j,x_0))\leq \epsilon$$
\noindent
Let $\epsilon>0$, and write $B_{x}:=Band(x_1,...,x_j,x)$ and $B_{x_0}:=Band(x_1,...,x_j,x_0)$ for the simplicity of notation. We have :
$$d_H(B_{x},B(x_0))= max\left(\underset{z\in B_x}{\text{sup }} d(z,B_{x_0}),\underset{z\in B_{x_0}}{\text{sup }} d(z,B_{x}) \right) $$
with $d$ being the distance induced by $||.||_{\infty}$.\\
It is easy to see that for any $z\in B_x$, $\underset{y \in B_{x_0}}{\text{inf }} ||z-y||_{\infty}$ is minimized by the function $y^{*}(t)=z(t)\mathds{1}_{(z(t) \in B_{x_0})}+max(x_1(t),...,x_j(t),x(t))\mathds{1}_{(z(t) \notin B_{x_0})}$. \\

Following this, we have :
 \begin{align*}
  d(z,B_{x_0})&= ||z-y^{*}||_{\infty}\\&=\max (  \underset{t : z(t) \notin B_{x_0},z(t)>B_{x_0}}{\text{sup }} |z(t)-max(x_1(t),...,x_j(t),x(t))|,\\&  \underset{t : z(t) \notin B_{x_0}, z(t)<B_{x_0}}{\text{sup }} |z(t)-min(x_1(t),...,x_j(t),x(t))| )
\end{align*}
 \noindent
 $z\in B_x$ implies that $\forall t, \quad   min\left( x(t),\underset{i=1,...,j}{\text{min }}X_i(t)\right)\leq z(t)\leq max\left( x(t),\underset{i=1,...,j}{\text{max }}X_i(t)\right)$ . If $z(t)>B_{x_0}$ too,  $max(\underset{i=1,...,j}{\text{max }}X_i(t),x_0(t))< z(t) \leq x(t)$ and $$  \underset{t : z(t) \notin B_{x_0},z(t)>B_{x_0}}{\text{sup }} |z(t)-max(x_1(t),...,x_j(t),x(t))| =\underset{t : z(t) \notin B_{x_0},z(t)>B_{x_0}}{\text{sup }}|z(t)-x(t)|.$$ With the same argument we have : $$  \underset{t : z(t) \notin B_{x_0},z(t)<B_{x_0}}{\text{sup }} |z(t)-max(x_1(t),...,x_j(t),x(t))| =\underset{t : z(t) \notin B_{x_0},z(t)<B_{x_0}}{\text{sup }}|z(t)-x(t)|.$$
It follows that for every $z\in B_x$ , $d(z,B_{x_0})\leq ||x-x_0 ||_{\infty}\leq \epsilon$ (simply by taking $\delta=\epsilon$). We then have $$\underset{z\in B_x}{\text{sup }} d(z,B_{x_0})\leq \epsilon\,.$$
We can prove that $$\underset{z\in B_{x_0}}{\text{sup }} d(z,B_{x}) \leq \epsilon$$ with the same argument which lead us to the final result.

 
 
\section{Monte-Carlo approximation of the average version of the empirical ACH depth.}\label{sec:algoappr}

The procedure given in Section 3.2 can be summarize by the Algorithm ~\ref{alg:main}.
\begin{algorithm}
    \label{alg:main}
    {\bf Input:} $\tilde{\mathcal{S}}_n=\{X_1',\ldots, X_n'\}$-dataset, the observed curve $x'$, $K$, $1 \leq J\leq n$ and the vector of weights $\mathbf{w}=(w_j)_{1\leq j\leq J} $ such that $\mathbf{w}_j=\frac{\binom{j}{n}}{\sum_{m=1}^{J} \binom{m}{n}}$.
    \begin{enumerate}
        \item For $k=1,\ldots,K$ do:
        \begin{itemize}
        \item[(i)]Select $ l \in \{1, \ldots, J \}$  according to $\mathbf{w}$.
        \vspace*{0.2cm}
        \item[(ii)] Select randomly and uniformly $(i_1,\ldots,i_l )\in \{1, \ldots, n \}$.
                \vspace*{0.2cm}
        \item[(iii)] $s(x') \leftarrow s(x')+ \frac{\lambda_{2}\left(conv \left(graph \left( \{X_{i_1}',\ldots,X_{i_l}' \}\right) \right) \right)}{\lambda_2 \left( conv \left( graph \left( \{X_{i_1}', \ldots,X_{i_{l}}', x'\}\right) \right) \right)}$.
        \end{itemize}
        
        \item {\bf Return:} $\quad \overline{D}_{J,n}(x'|\mathcal{S}_n') = \frac{1}{K}s(x')$.
    \end{enumerate}
\end{algorithm}

\section{Additional Experiences}\label{sec:addexper}

\subsection{Asymptotic variance of the exact and approximate versions}\label{subsec:var}

\begin{figure*}[!h]
\begin{center}
\begin{tabular}{cc}
$x_0$ & $x_1$\\
\includegraphics[height=.16\textheight, trim=.0cm .0cm 0cm 1cm,clip=true]{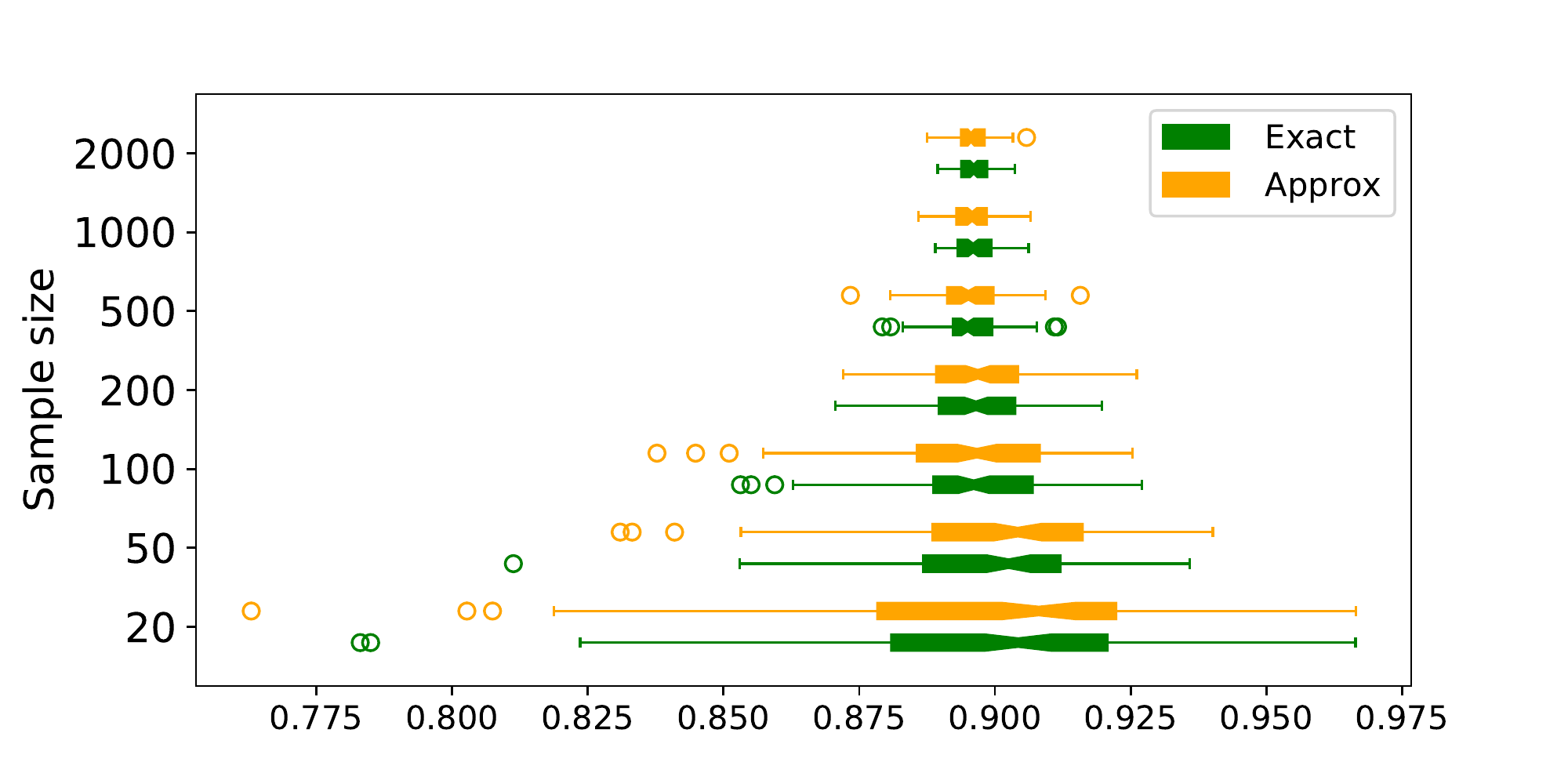}&\includegraphics[height=.16\textheight, trim=0cm .0cm 0cm 1cm,clip=true]{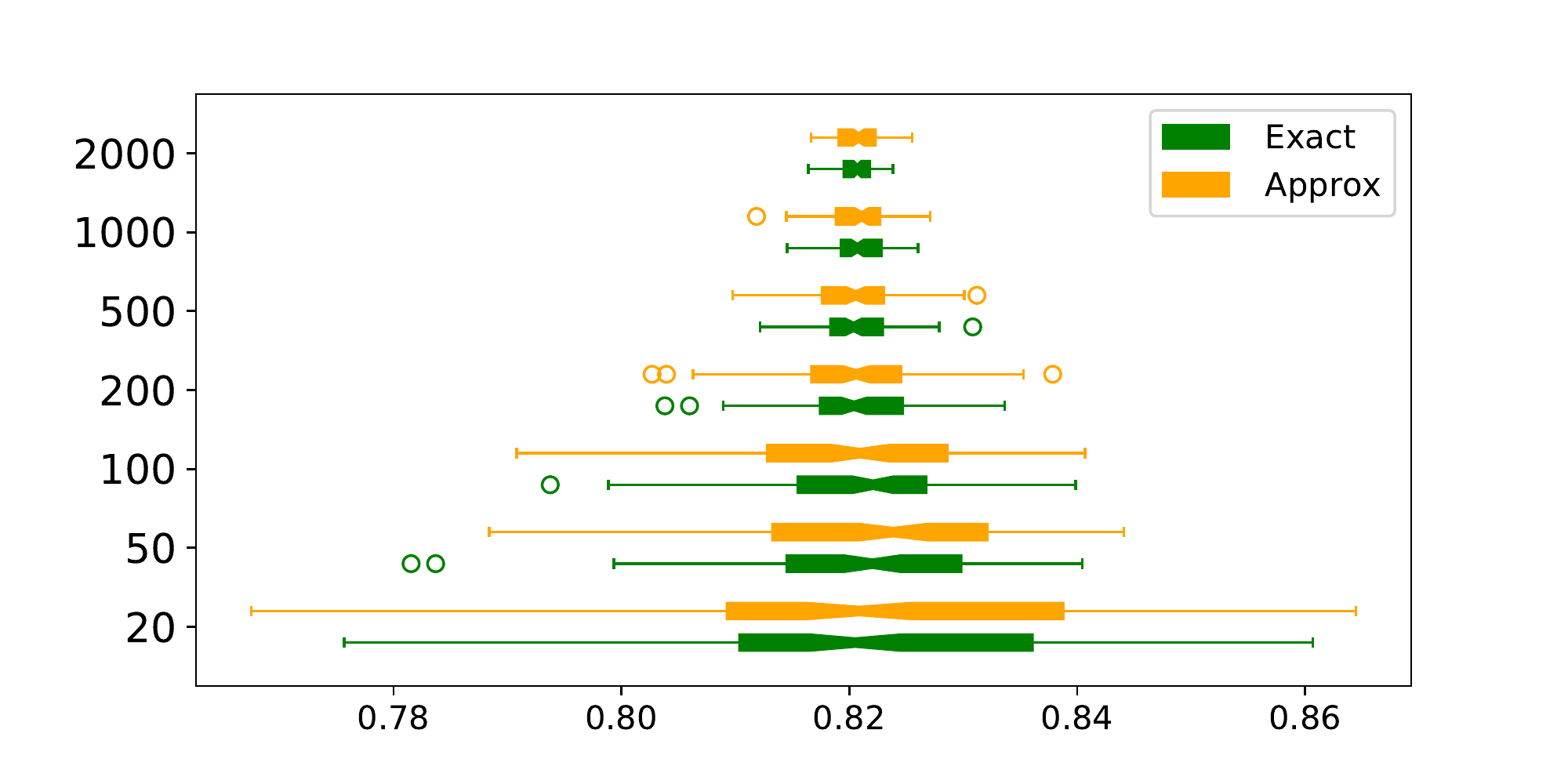}\\
$x_2$ & $x_3$\\
 \,\includegraphics[height=.16\textheight, trim=0cm 0cm 0cm 1cm,clip=true]{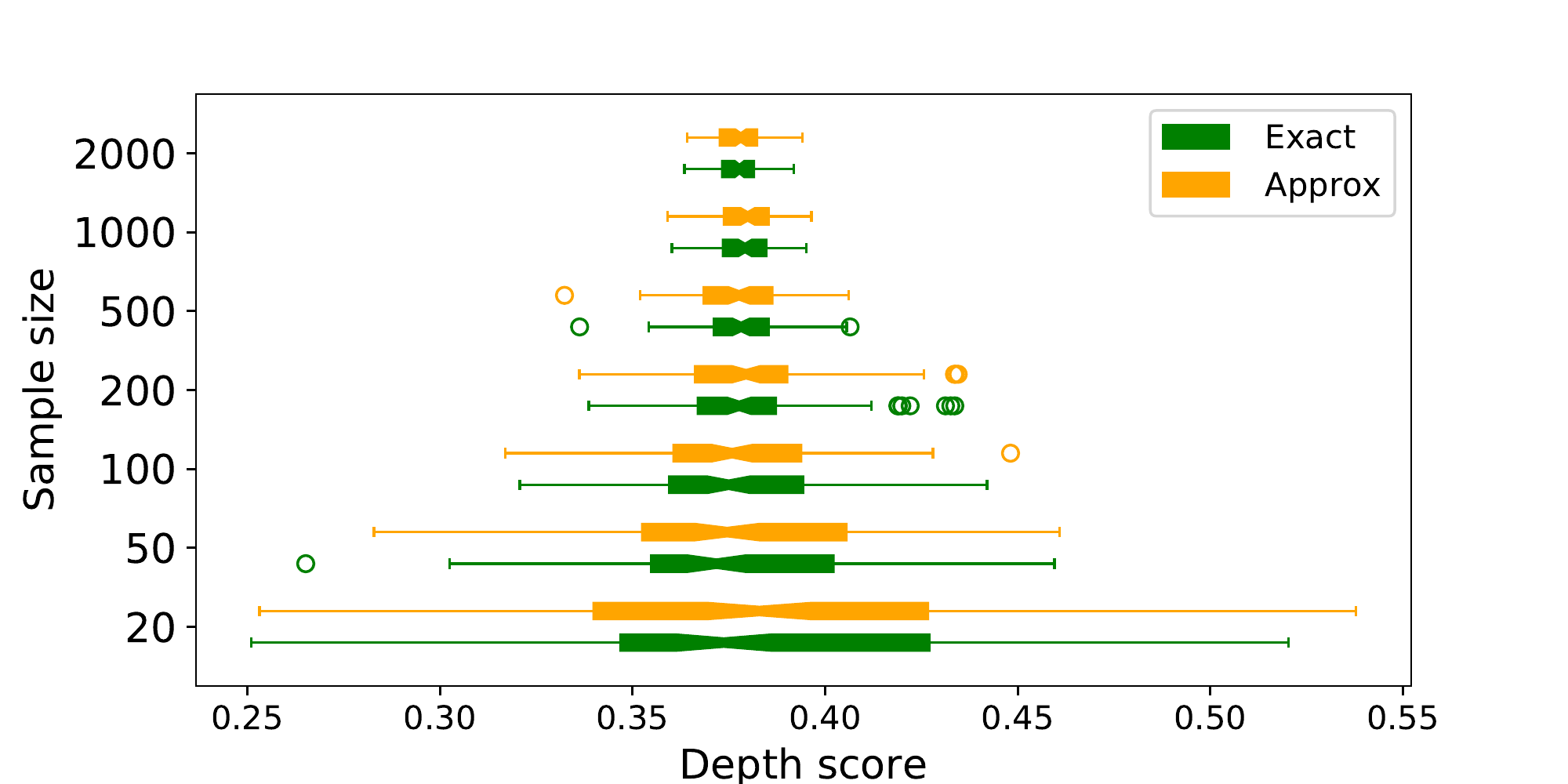}&\includegraphics[height=.16\textheight, trim=0cm .0cm 0cm 1cm,clip=true]{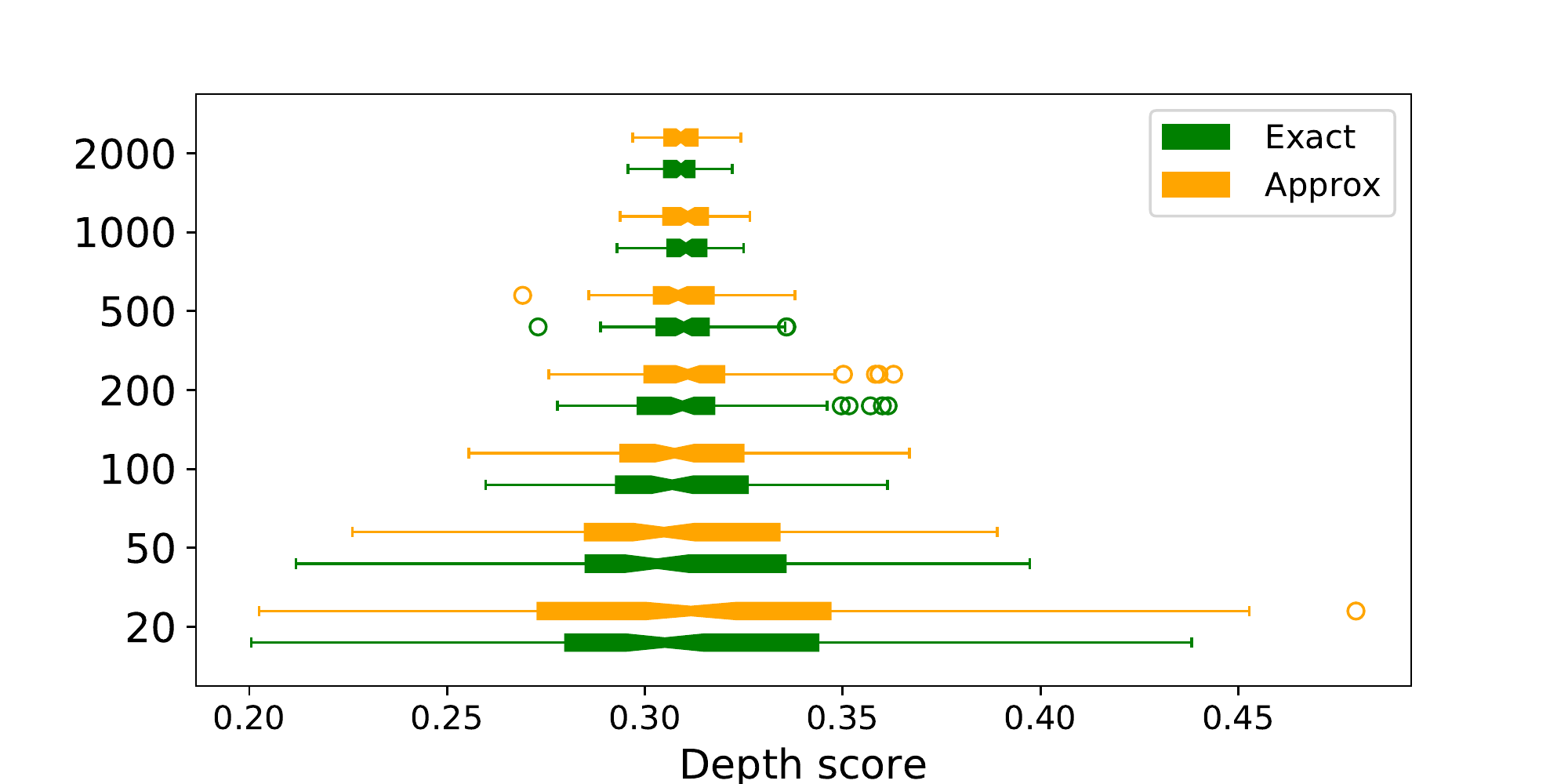}
 \end{tabular}
\end{center}
\caption{Boxplot (over $100$ repetitions) of the depth score for the observations $\mathbf{x}_0,\mathbf{x}_1,\mathbf{x}_2,\mathbf{x}_3$ for the two following settings on the data set (a): the green boxplots represent the exact computation while  the orange boxplots represent the approximation both with $J=2$.}
\label{fig:sampleD1}
\end{figure*}

To obtain further insights about the stability of the proposed depth notion, we explore its asymptotic variance. For this, we compute (exact and approximate) ACHD of $x_i,\,i=1,2,3,4$ for different sample sizes. The boxplots over $100$ repeated simulation for data sets (a) and (b) are indicated in Fig.~\ref{fig:sampleD1} and ~\ref{fig:sampleD2}. One observes not only stable decrease of the variance of ACHD with the sample size, but also the similarity between exact and approximate versions, which hints on stability and precision of the exact algorithm even when exploring a small portion of combinations (e.g., when $n=500$ only $2\%$ of all pairs are explored for $K=5n$).

\begin{figure*}[h]
\begin{center}
\begin{tabular}{cc}
$x_0$ & $x_1$\\
\includegraphics[height=.16\textheight, trim=.0cm .0cm 0cm 1cm,clip=true]{boxplot_variance_D1_x0.pdf}&\includegraphics[height=.16\textheight, trim=0cm .0cm 0cm 1cm,clip=true]{boxplot_variance_D1_x1.pdf}\\
$x_2$ & $x_3$\\
 \,\includegraphics[height=.16\textheight, trim=0cm 0cm 0cm 1cm,clip=true]{boxplot_variance_D1_x2.pdf}&\includegraphics[height=.16\textheight, trim=0cm .0cm 0cm 1cm,clip=true]{boxplot_variance_D1_x3.pdf}
 \end{tabular}
\end{center}
\caption{Boxplot (over $100$ repetitions) of the depth score for the observations $\mathbf{x}_0,\mathbf{x}_1,\mathbf{x}_2,\mathbf{x}_3$ for the two following settings on the data set (b): the green boxplots represent the exact computation while  the orange boxplots represent the approximation both with $J=2$.}
\label{fig:sampleD2}
\end{figure*} 

\begin{figure}[!h]
\begin{center}
\includegraphics[scale=0.35]{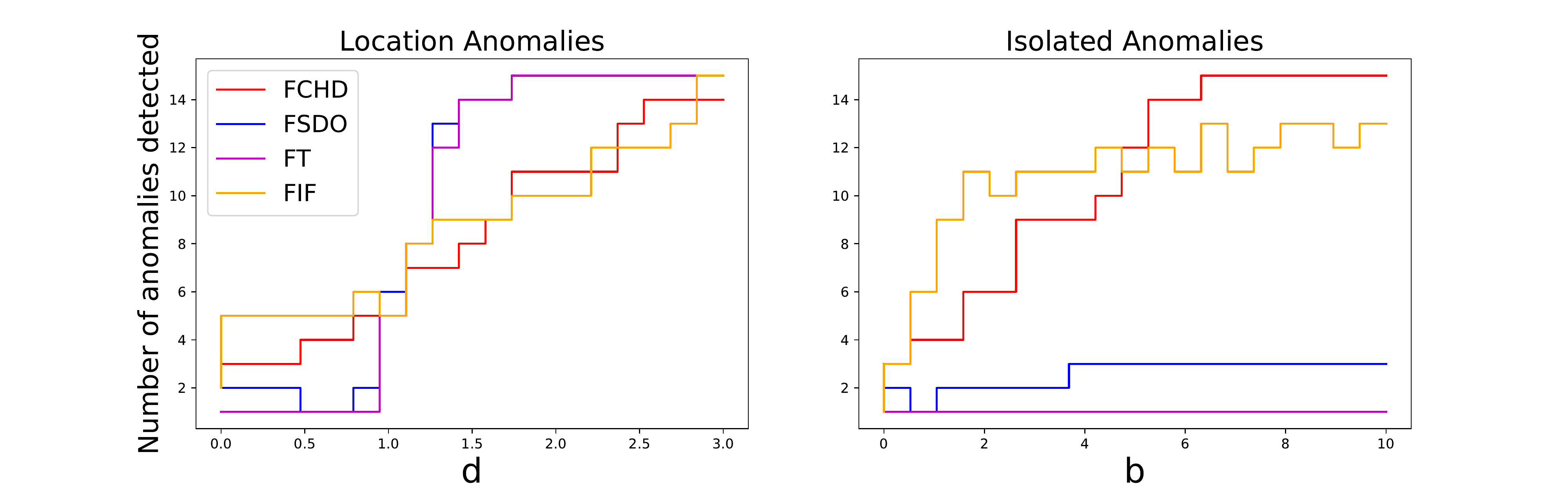}
\end{center}
\caption{Number of anomalies detected over a grid of parameters for two types of anomalies, location and isolated anomalies for ACHD and three others state-of-art methods.}
\label{anom2}
\end{figure}

\begin{figure}[!h]
\begin{tabular}{ccc}
\includegraphics[scale=0.35]{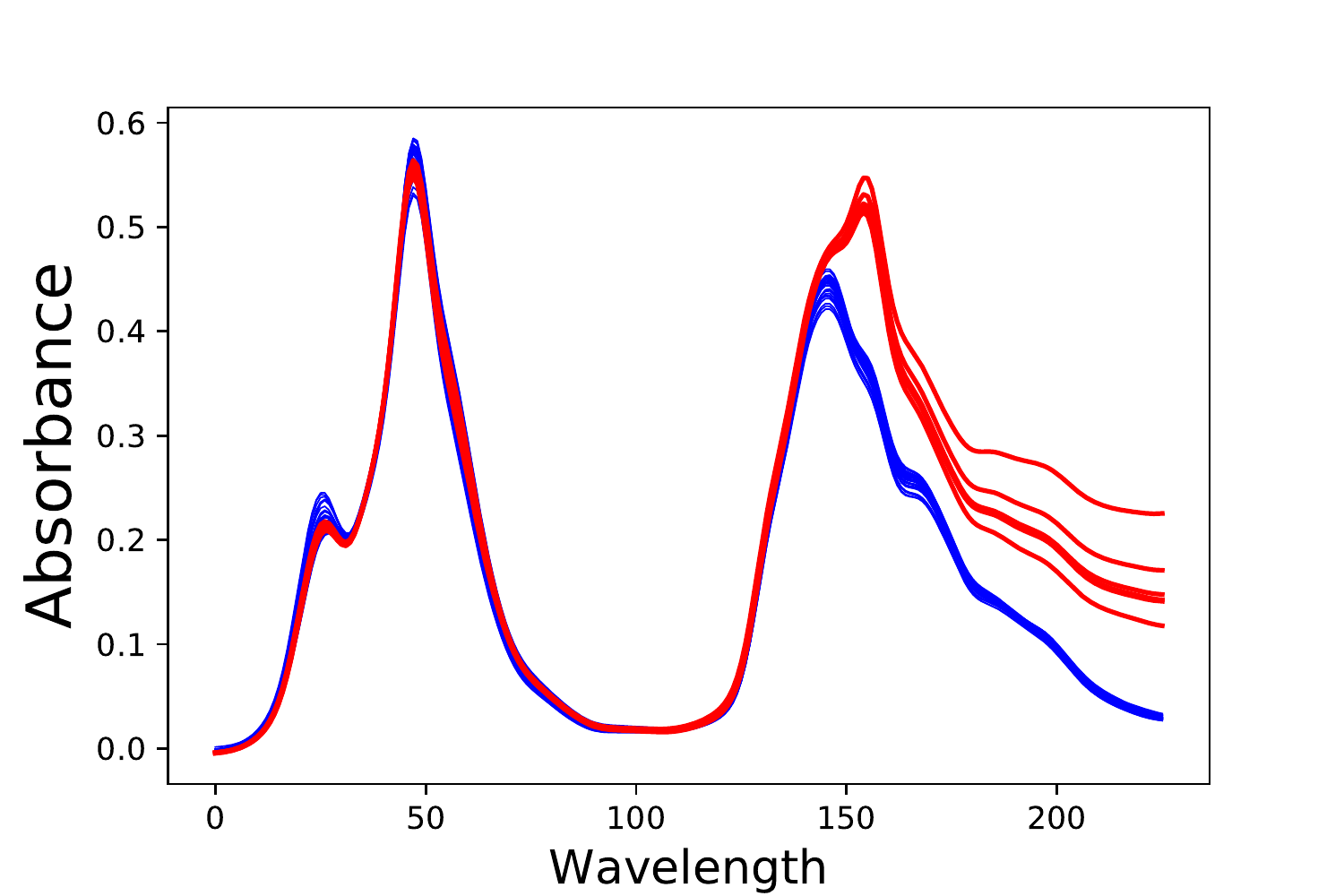}&\includegraphics[scale=0.35]{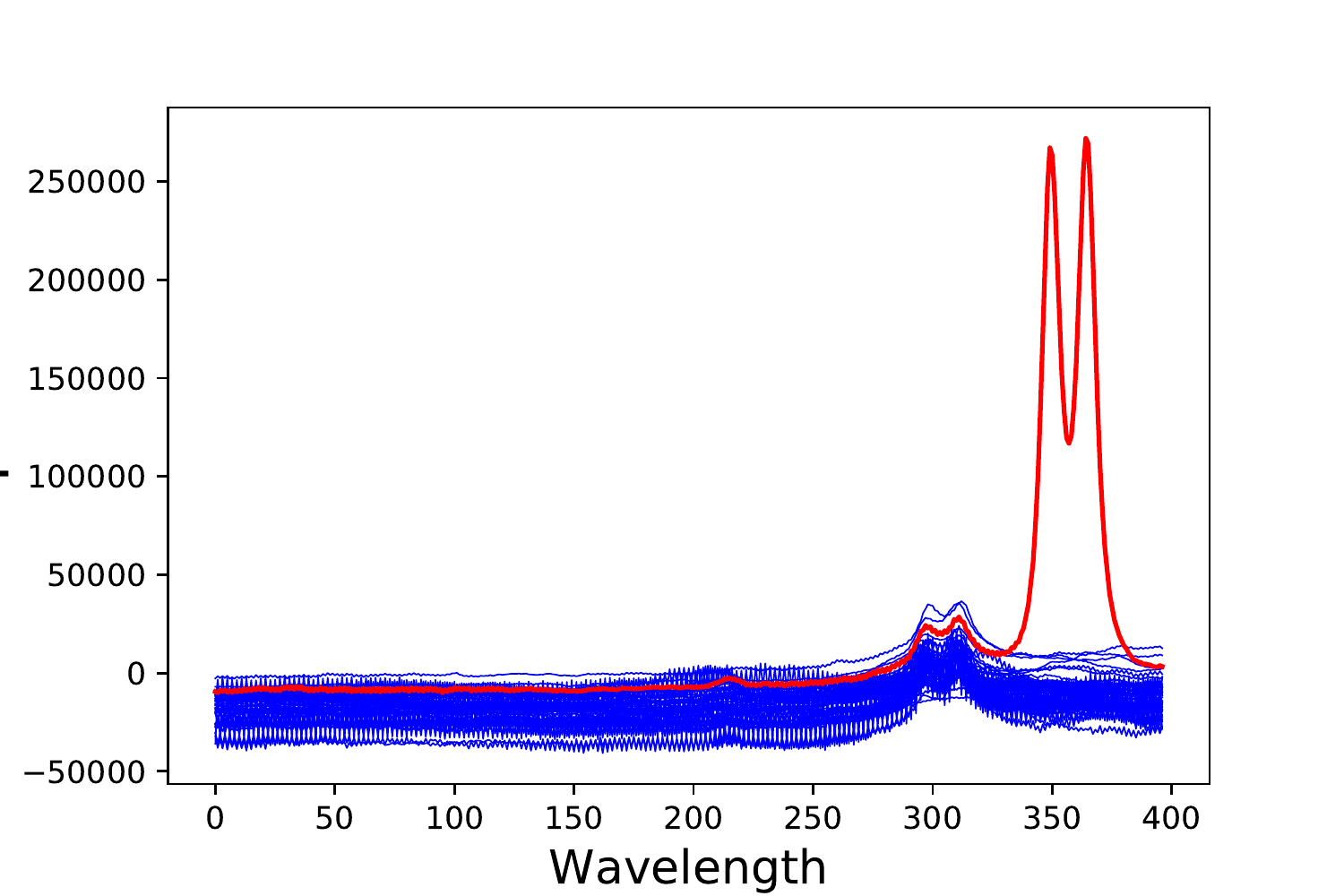}&\includegraphics[scale=0.35]{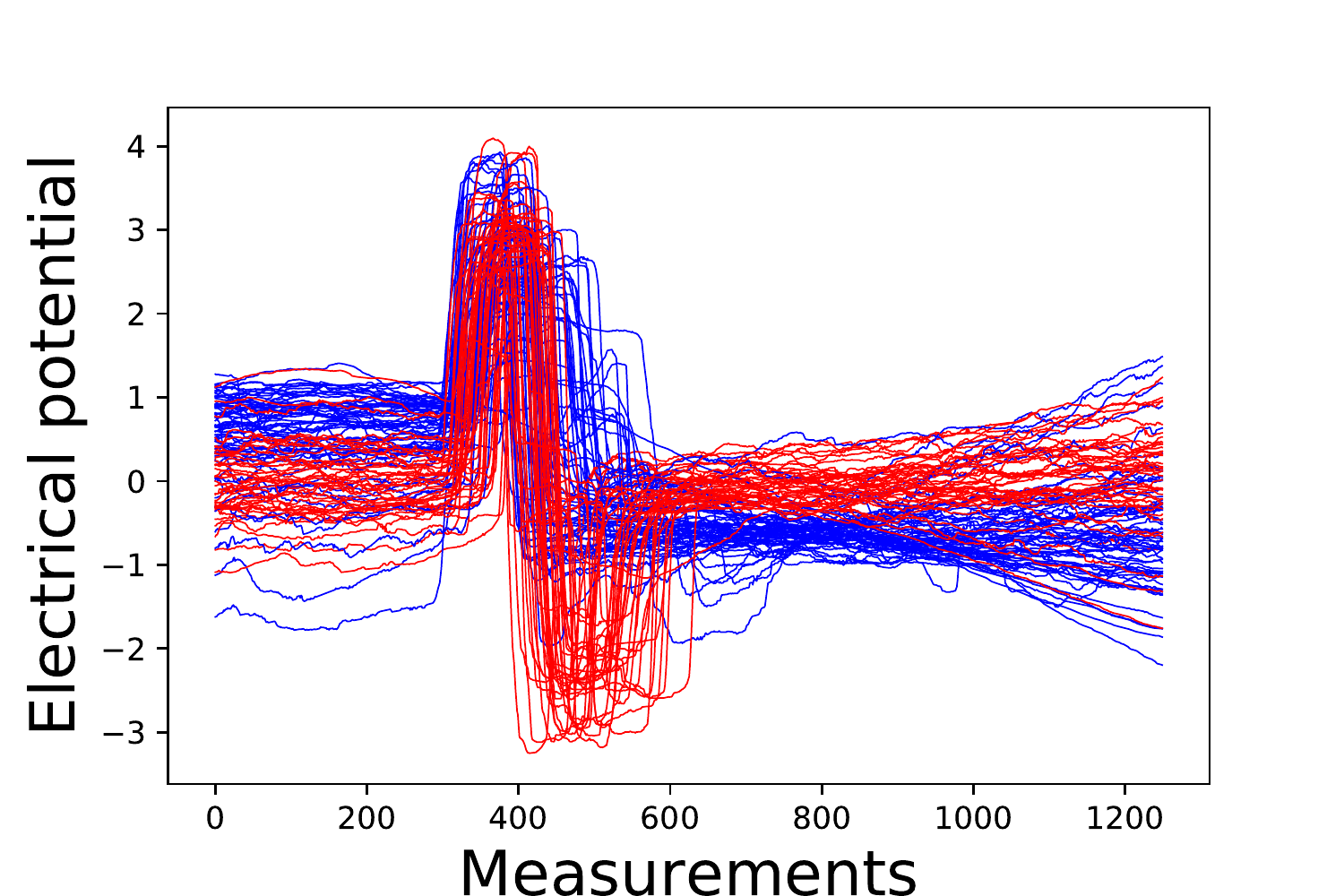}
\end{tabular}
\caption{The data sets Octane, Wine, and EOG used in Section~4.5.}
\label{data}
\end{figure}

The dataset Octane, Wine, and EOG used in the Section~4.4 are represented in Figure~\ref{data}.

\subsection{Robustness}

Additional results on the robustness experiment are given in Table ~\ref{rank} for the location anomalies with the dataset (b) and the isolated anomalies for the data set (a).

{\renewcommand{\arraystretch}{1} 
\begin{table*}[!h]
\begin{center}
\begin{tabular}{ |p{2cm}|  p{2cm}||  p{0.5cm}| p{1cm}|  p{1cm}|p{1cm}| p{1cm}|p{1cm}| }
 \hline
 \multicolumn{8}{|c|}{$d_{\tau}(\sigma_0, \sigma_{\alpha}) ( \times 10^{-2}) $ } \\
 \hline
& $\boldsymbol{ \alpha }$ & 0& 5 & 10 & 15& 25 & 30 \\
 \hline
 \hline
\multirow{2}{*}{ACHD} &Location& 0 & 0.7 &  1.5 &  2.3 &  4.2 &  5.1 \\
& Isolated & 0 & 1.3 & 1.8 & 1.6 & 2.4 & 3.2\\
\hline
\multirow{2}{*}{FSDO} &Location& 0   & 1.5 &3.1 & 5.1 & 8.8 & 11 \\
& Isolated &0   & 0.9 & 1 & 1.1 & 1.5 & 1.6 \\
 \hline
\multirow{2}{*}{FT} &Location&0 &  0.6 & 1.5 & 3 & 6.4 & 8  \\
& Isolated & 0 &1.3 &  0.8 & 0.9 & 1.1 & 1.5\\
 \hline
\multirow{2}{*}{FIF} &Location&0  & 14 & 15 & 15 & 16 & 15 \\
& Isolated & 0 & 6.9 & 7.3 & 7 & 8.2 & 8.1   \\
 \hline
\end{tabular}
\end{center}
\caption{Kendall's tau distances between the rank returned with normal data and contaminated data (over different size of contamination with location and isolated anomalies) for the area of the convex hull depth measure and three others state-of-art methods.}
\label{rank}
\end{table*}
}

\subsection{Anomaly detection}

Additional experiments with location anomalies from the data set (b) and isolated anomalies from data set (a) are given in Figure~\ref{anom2}.



\end{document}